\documentclass[pdflatex,sn-mathphys,iicol]{sn-jnl}
\usepackage{
graphicx,
amsfonts,
textcomp,
program,
amssymb,
listings,
rotating,
booktabs,
algorithmicx,
mathrsfs,
url,
appendix,
etoolbox,
algpseudocode,
multirow,
algorithm,
amsmath,
hyperref,
manyfoot,
hhline,
pifont}

\usepackage[utf8]{inputenc} 
\usepackage[T1]{fontenc}    
\usepackage{hyperref}       
\usepackage{url}            
\usepackage{booktabs}       
\usepackage{amsfonts}       
\usepackage{nicefrac}       
\usepackage{microtype}      
\usepackage{placeins}
\usepackage{colortbl} 
\usepackage{nicefrac}
\usepackage{dsfont}
\usepackage{marvosym}
\usepackage{fdsymbol}
\usepackage{graphicx}
\usepackage{adjustbox}

\usepackage{overpic}
\usepackage{makecell}
\usepackage{wrapfig}
\usepackage{marvosym}
\usepackage{caption}
\usepackage{colortbl}
\usepackage{xcolor}
\usepackage{nicefrac}
\usepackage{dsfont}
\usepackage{fdsymbol}
\usepackage{amssymb}
\usepackage{amsmath}
\usepackage{balance}

\usepackage{ragged2e}
\definecolor{OursColor}{HTML}{FFFFCC}  
\definecolor{myIDBcolor}{HTML}{FFF5F0}
\definecolor{myCCDBcolor}{HTML}{F5FFF0}  

\newcommand\shline{\specialrule{0.8pt}{0pt}{0pt}}

\newcommand{\best}[1]{\textbf{\textcolor{red}{#1}}}
\newcommand{\secondbest}[1]{\underline{\textcolor{blue}{#1}}}

\begin{document}

\title[Article Title]{\texorpdfstring{\raisebox{-5pt}{\includegraphics[height=30pt,width=28pt]{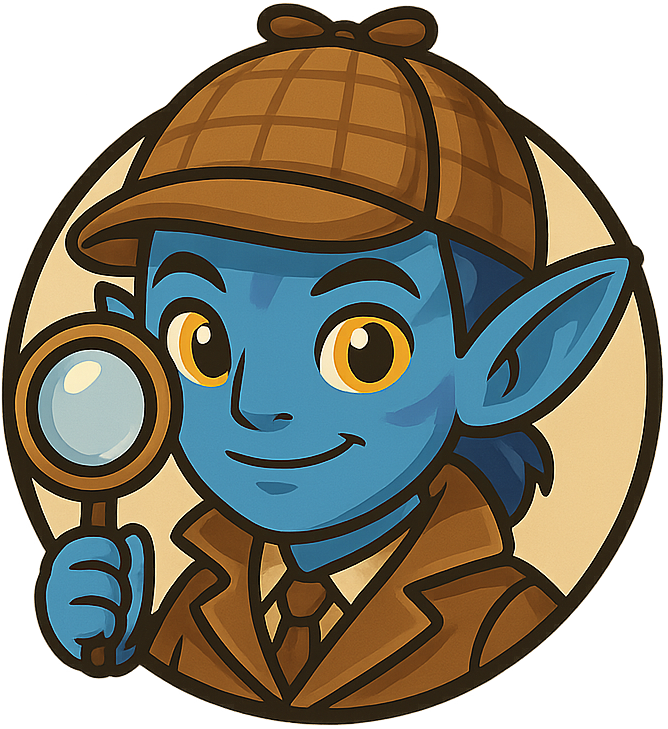}}}{Icon}AvatarShield: Visual Reinforcement Learning for Human-Centric Synthetic Video Detection}









\author[1]{\fnm{Zhipei} \sur{Xu}}\email{zhipeixu@stu.pku.edu.cn}
\equalcont{These authors contributed equally to this work.}

\author[1]{\fnm{Xuanyu} \sur{Zhang}}\email{xuanyuzhang21@stu.pku.edu.cn}
\equalcont{These authors contributed equally to this work.}

\author[1,2]{\fnm{Qing} \sur{Huang}}\email{huangqing011222@gmail.com}

\author[3]{\fnm{Xing} \sur{Zhou}}\email{zhouxing@tuzhanai.com}

\author*[1]{\fnm{Jian} \sur{Zhang}}\email{zhangjian.sz@pku.edu.cn}

\affil[1]{\orgdiv{School of Electronic and
Computer Engineering, Peking University}, 
\orgaddress{\city{Shenzhen}, \country{China}}}
\affil[2]{\orgdiv{School of Future Technology, South China University of Technology}, 
\orgaddress{\city{Guangzhou}, \country{China}}}
\affil[3]{\orgdiv{RabbitPre AI}, 
\orgaddress{\city{Shenzhen}, \country{China}}}


\abstract{Recent advances in Artificial Intelligence Generated Content have led to highly realistic synthetic videos, particularly in human-centric scenarios involving speech, gestures, and full-body motion, posing serious threats to information authenticity and public trust. Unlike DeepFake techniques that focus on localized facial manipulation, human-centric video generation methods can synthesize entire human bodies with controllable movements, enabling complex interactions with environments, objects, and even other people. However, existing detection methods largely overlook the growing risks posed by such full-body synthetic content. Meanwhile, a growing body of research has explored leveraging LLMs for interpretable fake detection, aiming to explain decisions in natural language. Yet these approaches heavily depend on supervised fine-tuning, which introduces limitations such as annotation bias, hallucinated supervision, and weakened generalization. To address these challenges, we propose AvatarShield, a novel multimodal human-centric synthetic video detection framework that eliminates the need for dense textual supervision by adopting Group Relative Policy Optimization, enabling LLMs to develop reasoning capabilities from simple binary labels. Our architecture combines a discrete vision tower for high-level semantic inconsistencies and a residual extractor for fine-grained artifact analysis. We further introduce FakeHumanVid, a large-scale benchmark containing 15K real and synthetic videos across nine state-of-the-art human generation methods driven by text, pose, or audio. Extensive experiments demonstrate that AvatarShield outperforms existing methods in both in-domain and cross-domain settings.
}


\maketitle

\section{Introduction}

Recently, Artificial Intelligence Generated Content (AIGC) technologies have developed rapidly, achieving remarkable progress especially in the field of video generation. Advanced video generative models such as Sora~\cite{brooks2024video}, Kling~\cite{Klingai2024}, and stable diffusion video~\cite{blattmann2023stable} have demonstrated outstanding capabilities in creating realistic videos, significantly enhancing creative efficiency, but simultaneously posing serious challenges to information authenticity. 
Among these advancements, human-centric video generation~\cite{cui2024hallo3, tu2024stableanimator, zhang2024mimicmotion} has emerged as a particularly impactful category. Unlike DeepFake techniques, which are typically limited to localized facial manipulation, human-centric generation methods are capable of synthesizing videos of entire human figures, including facial expressions, body movements, and interactions with surrounding environments, objects, and other individuals. Powered by multimodal conditioning inputs such as pose, audio, and text, these methods support high degrees of controllability and realism. 
In terms of societal harm, human-centric synthetic videos represent a significantly greater threat: in contrast to DeepFakes, which are often limited to face-level impersonation, human-centric synthetic videos can fabricate entirely fictitious human actions, behaviors, and social interactions, making them far more effective for orchestrating large-scale deception, manipulating public opinion, or fabricating plausible yet non-existent events.
As a result, human-centric synthetic videos have blurred the boundaries between artificial and genuine content, carrying heightened risks, making their detection significantly more urgent and challenging.

Current state-of-the-art video detection methods~\cite{chen2024demamba,song2024learning,kong2024open,kong2024moe,luo2023beyond} have achieved remarkable performance via carefully constructed large-scale datasets and complex designs. However, several critical issues remain unresolved. 

\textbf{(I) Threats from Human-Centric Generation Video:}
Most existing AI-generated video benchmarks~\cite{chen2024demamba} and detection approaches~\cite{song2024learning} focus on general scenarios such as natural landscapes, animals, plants, or cartoon characters, where distinguishing between real and synthetic content is relatively easy or less consequential. These videos are typically created for entertainment purposes and are unlikely to trigger serious societal crises or public opinion risks.
More importantly, some human-centric generation techniques such as lip-synchronization~\cite{cui2024hallo3} or pose-driven video generation~\cite{tu2024stableanimator} achieve more realistic results by introducing stronger conditional controls during the generation process. Because these videos involve human speech and actions, they are more likely to lead to violations of portrait rights and spark related legal disputes.
Nevertheless, existing research lacks dedicated methods and benchmark datasets explicitly tailored to the detection of human-centric video generation.


\textbf{(II) Limitations of Supervised Fine-Tuning (SFT):}
To break the black-box nature of traditional forgery detection methods, some recent works~\cite{xu2024fakeshield,huang2024ffaa,liu2024forgerygpt} have introduced large language models (LLMs), aiming to present the reasoning behind detection decisions in textual form. However, these approaches heavily rely on SFT to optimize the LLMs, which introduces several limitations.
\textbf{First}, SFT depends on densely annotated image-text datasets, which are typically generated either through automated annotation using LLM tools like GPT-4o~\cite{achiam2023gpt} or via manual labeling by human experts. However, the former approach is prone to hallucinations, while the latter is susceptible to subjective bias. Both approaches struggle to guarantee high-quality supervision, potentially capping the model’s performance. Additionally, the annotation process itself is time-consuming and labor-intensive.
\textbf{Second}, studies~\cite{chu2025sft} have shown that SFT can sometimes lead to rote memorization, whereas reinforcement learning (RL) tends to foster stronger generalization capabilities. We argue that while SFT helps models acquire detection capabilities, it compromises the general question-answering and visual perception abilities of LLMs.
In fact, LLMs inherently possess a certain ability to assess the authenticity of visual content~\cite{song2024learning}. Rather than relying on ``\textbf{spoon-fed teaching methods}'', we aim to \textbf{inspire and enhance} the LLMs' innate reasoning and perception capabilities for more effective fake detection.

\begin{figure*}[t!]
	\centering
    \includegraphics[width=1.0\linewidth]{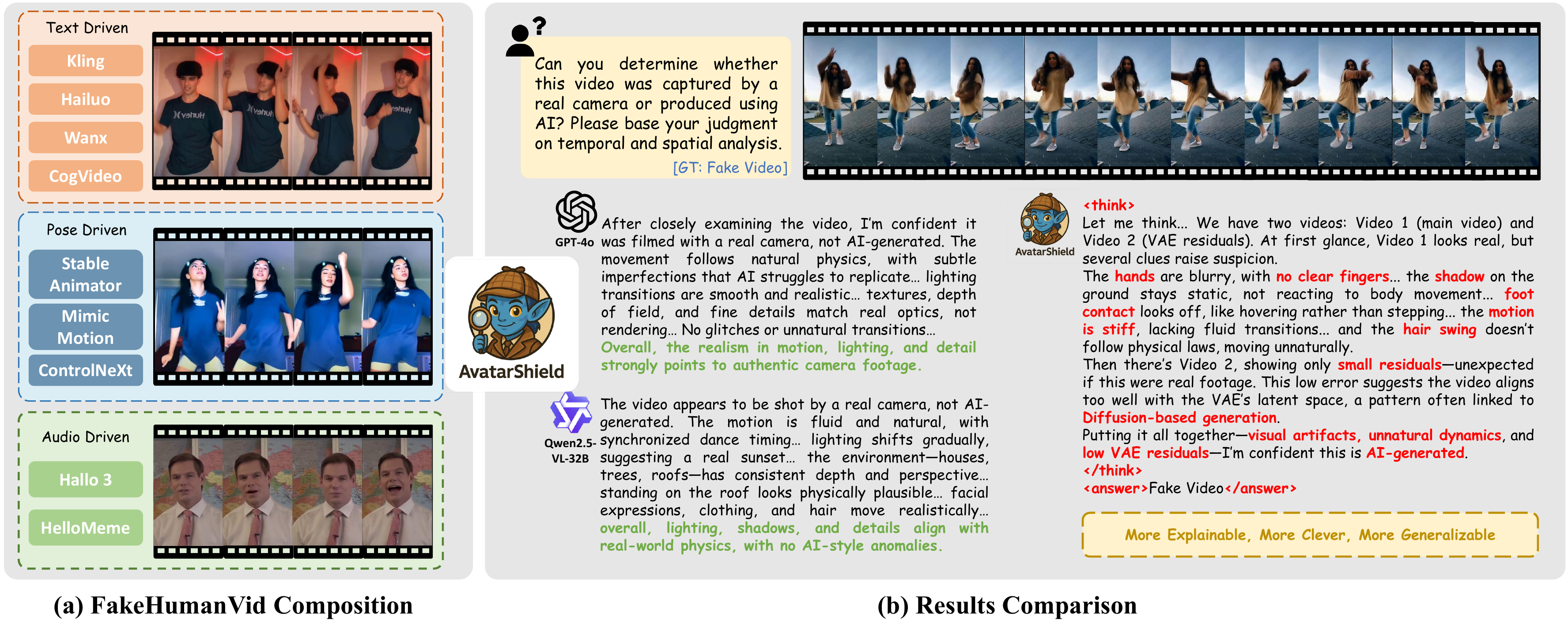}
	\vspace{-5pt}
	\caption{We focus on the human-centric generation video detection problem, constructing a human synthetic detection dataset FakeHumanVid, along with an efficient reasoning-style multi-modal large language model AvatarShield. Our FakeHumanVid dataset encompasses 9 different pose-driven, text-driven, and audio-driven video generation methods. The proposed AdavarShield significantly outperforms existing mainstream LLMs in terms of detection accuracy and reasoning capabilities.}
	\label{teasor}
    \vspace{-16pt}
\end{figure*}

Reinforcement learning has recently demonstrated strong capability in guiding and optimizing LLMs, especially with the introduction of Group Relative Policy Optimization (GRPO)~\cite{guo2025deepseek}, which has attracted significant attention. GRPO’s main innovation lies in replacing the explicit value model with group-wise comparison, lowering computational costs. Compared to RLHF~\cite{ouyang2022training}, GRPO is outcome-driven and does not require annotations of the reasoning process, which greatly reduces the need for labeled textual data. Recent studies~\cite{zhao2025r1,feng2025video,shen2025vlm, chen2025r1v, tan2025reason,khalid2021fakeavceleb} have also extended GRPO to multimodal large language models (MLLMs), showing strong performance and great potential for generalization across various tasks such as visual reasoning, medical image analysis, and OCR. 
In the field of synthetic detection, where new generation techniques are rapidly emerging, enhancing generalization and zero-shot detection capabilities is of paramount importance. Existing studies~\cite{chu2025sft} have demonstrated that RL, compared to SFT, is more effective in improving the generalization ability of LLMs on unseen data.
Moreover, GRPO requires only minimal supervision, such as simple labels like ``\textit{Fake Video}'' or ``\textit{Real Video}'', to prompt the LLM to engage in autonomous reasoning. Interestingly, the resulting thinking process closely resembles the explanatory reasoning texts found in previous SFT-constructed datasets~\cite{xu2024fakeshield,kang2025legion,narayan2023df,cai2024av}, offering interpretability for the detection outcomes.
Therefore, we explore the feasibility of using GRPO to optimize LLMs, aiming to eliminate the need for costly manual video annotation while improving generalization to unseen synthetic types.

To address the two major challenges in current synthetic video detection methods, we present a novel explainable human-centric synthetic video detection framework, \textbf{AvatarShield}, along with a new benchmark dataset, \textbf{FakeHumanVid}. As illustrated in Figure~\ref{teasor}, we categorize state-of-the-art human video generation methods into three types based on their conditioning inputs: pose-driven, audio-driven, and text-driven. We construct massive fake videos using methods such as StableAnimator~\cite{tu2024stableanimator}, Kling~\cite{Klingai2024}, and Hallo3~\cite{cui2024hallo3}, and collect real videos from public datasets~\cite{zhang2021hdtf,Jafarian_2021_CVPR_TikTok}, jointly forming the FakeHumanVid benchmark. Leveraging this dataset, we apply the GRPO algorithm to optimize LLMs and enhance their ability to identify fake content. 
Furthermore, we introduce a dual-encoder architecture to capture both high-level and low-level anomalies. The high-level encoder uses a discrete vision tower~\cite{Qwen2.5-VL} to encode spatial domain video features, enabling the LLM to perceive semantic inconsistencies across temporal frames. The low-level encoder employs VQ-VAE~\cite{van2017neural} to amplify generation artifacts, allowing the LLM to focus on fine-grained visual inconsistencies. Our main contributions are summarized as follows:



\vspace{1pt}
\noindent \ding{113}~(1) We present a novel pure LLM-based, reasoning-style framework for detecting human-centric synthetic videos in an end-to-end manner. It requires only real/fake labels to guide the model toward generalizable and interpretable detection reasoning, significantly reducing the need for extensive textual annotations required by SFT.

\vspace{1pt}
\noindent \ding{113}~(2) We introduce a dual-encoder architecture that combines a discrete vision tower to extract high-level semantic features and a continuous VQ-VAE to amplify low-level generation artifacts. Furthermore, we design the accuracy reward and temporal compensation reward, enabling the network to achieve high detection accuracy and strong temporal modeling capabilities.

\vspace{1pt}
\noindent \ding{113}~(3) Focusing on the human-centric generation video detection, we build a comprehensive dataset, FakeHumanVid, covering 9 major synthetic methods. Our FakeHumanVid contains 13.5k videos for training and 1.5k for testing, effectively serving as a benchmark to evaluate human-centric generation video detectors in real-world scenarios.

\vspace{1pt}
\noindent \ding{113}~(4) Extensive experiments on the FakeHumanVid benchmark demonstrate that our method outperforms all mainstream fake detection methods in both cross-domain and in-domain settings.




\section{Related Works}

\subsection{Non-LLM-Based AI Forgery Detection}



Traditional AI forgery detection methods largely revolve around using CNN-based or transformer-based architectures tailored to the task of identifying synthetic visual content~\cite{zhong2023patchcraft,zhong2023rich,zhang2022patch,yu2024semgir,fang2024uniforensics,salvi2023robust,tan2025c2p,li2025texture,zhu2024hiding,yan2024effort,yang2025all,yan2024generalizing,zhou2024freqblender,luo2024forgery,guo2025rethinking,narayan2025facexbench,cui2024forensics,li2020face,huang2023implicit,yan2023ucf}. These approaches focus on pixel-level, frequency-domain, or statistical inconsistencies within tampered images or videos. For instance, CNNSpot~\cite{wang2019cnnspot} employed strategic data augmentation and robust convolutional structures to detect GAN-generated samples. FreDect~\cite{frank2020fredect} operated in the frequency domain to expose upsampling artifacts typical in synthetic images. DIRE~\cite{wang2023dire} took a different route by utilizing reconstruction errors derived from diffusion models, amplifying a significant gap between real and generated image distributions. 
AIDE~\cite{yan2024sanity} incorporated global-aware local feature disentanglement to handle domain shifts in fake image detection. Uni-FD~\cite{ojha2023fakedetect} introduced a CLIP-based nearest neighbor method, demonstrating impressive generalization to unseen generative models using a frozen vision-language model’s feature space.
CFM~\cite{luo2023beyond} enhanced model generalization and robustness by combining prior-agnostic data augmentation with fine-grained relation learning and a progressive controller to mine and focus on critical forgery features.
LSDA~\cite{yan2024transcending} tackled the generalization challenge by augmenting forgery diversity in the latent space, enabling the model to learn smoother transitions and more generalizable decision boundaries across various forgery types.
RECCE~\cite{cao2022end} approached forgery detection by learning compact representations of genuine faces via joint reconstruction-classification learning, using multi-scale bipartite graphs and reconstruction-guided features to generalize beyond known forgery patterns.
UCF~\cite{yan2023ucf} addressed both forgery-irrelevant and method-specific overfitting by disentangling image features into three components, leveraging multi-task learning and contrastive regularization to isolate common forgery cues for improved generalization.
IID~\cite{huang2023implicit} approached face swapping detection by exploiting both explicit and implicit face identities, using explicit identity contrast (EIC) and implicit identity exploration (IIE) losses to enhance the distinction between real and fake faces.
However, these non-LLM based methods operate as black-box classifiers, lacking interpretable or stepwise reasoning. This absence of transparency hinders understanding of their decision-making process and limits their trustworthiness, especially in critical applications.

\begin{figure*}[t!]
	\centering
    \includegraphics[width=1.0\linewidth]{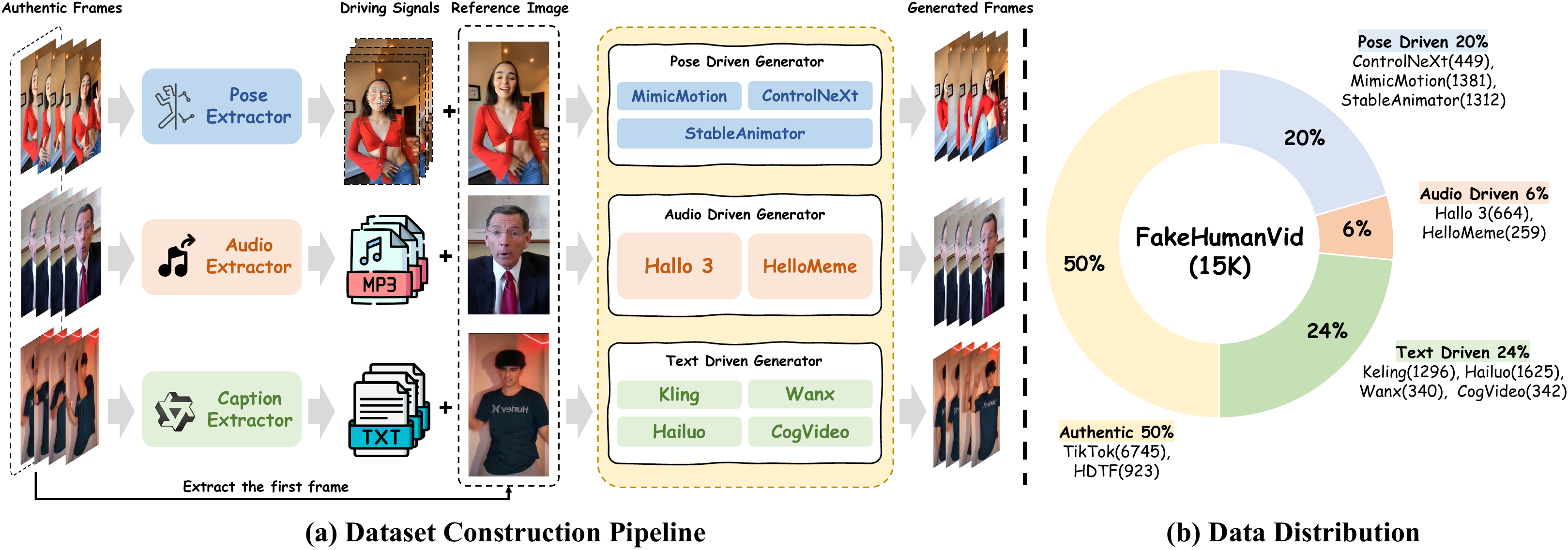}
	\vspace{-10pt}
	\caption{Construction process and data distribution of our proposed FakeHumanVid.}
	\label{dataset}
    \vspace{-16pt}
\end{figure*}

\subsection{LLM-based AI Forgery Detection}

Recent advances in LLM-based forgery detection~\cite{huang2024ffaa,yan2025gpt,chen2024textit,yan2024df40,yan2024transcending} have led to more generalizable and interpretable methods, with the reasoning and understanding capabilities. Early approaches such as FakeShield~\cite{xu2024fakeshield}, which combines large language models with image encoders to detect and explain image forgeries via pixel-level artifacts and semantic inconsistencies, and ForgeryGPT~\cite{liu2024forgerygpt}, which integrates forgery detection into a language model using a mask-aware forgery extractor and multi-stage training, highlight the trend of multimodal solutions. Other methods include SIDA~\cite{huang2024sida}, which targets social media deepfakes with a specialized multimodal model, and LEGION~\cite{kang2025legion}, which enhances interpretability through artifact localization and textual explanation.
However, most of these approaches have focused primarily on image-based forgery detection. While MM-Det~\cite{song2024learning} recently extended detection capabilities to AI-generated videos, it mainly targeted general scene understanding rather than human-centric content (such as faces or behaviors). Moreover, it is limited to explaining image-level anomalies and lacks the capacity for coherent video-level understanding. Despite these advancements, current LLM-based detection methods still face several limitations. They rely heavily on SFT using large-scale annotated datasets that include explicit textual reasoning processes. Such training procedures often encourage superficial pattern recognition rather than genuine heuristic or reasoning-based understanding. Additionally, there has been minimal exploration of purely LLM-driven pipelines.

\section{Methodology}
\label{Methodology}

\subsection{Construction of FakeHumanVid}
\label{Construction of FakeHumanVid}

\textbf{Motivation:} Existing synthetic video datasets primarily focus on broad scenarios such as natural scenes and animated content. However, from the perspective of societal harm and misinformation propagation, human-centric generation videos pose a significantly greater risk. Furthermore, current datasets mostly include general video generation methods, such as Sora~\cite{brooks2024video} and Kling~\cite{Klingai2024}, while overlooking specialized human-focused generation methods like pose-driven dance synthesis~\cite{tu2024stableanimator} or audio-driven speaker generation~\cite{cui2024hallo3}. These techniques can often generate highly realistic and convincing human videos, making detection significantly more challenging.
Motivated by the work of~\cite{lei2024comprehensive}, we categorize human-centric video generation methods based on their guiding conditions into three groups: pose-driven, audio-driven, and text-driven. We select 9 representative generation methods~\cite{zhang2024mimicmotion,peng2024controlnext,tu2024stableanimator,cui2024hallo3,zhang2024hellomeme,Klingai2024,hailuo,yang2024cogvideox,wang2025wan}, with the detailed dataset construction pipeline and distribution illustrated in Figure~\ref{dataset}.

\begin{figure*}[t!]
	\centering
    \includegraphics[width=1\linewidth]{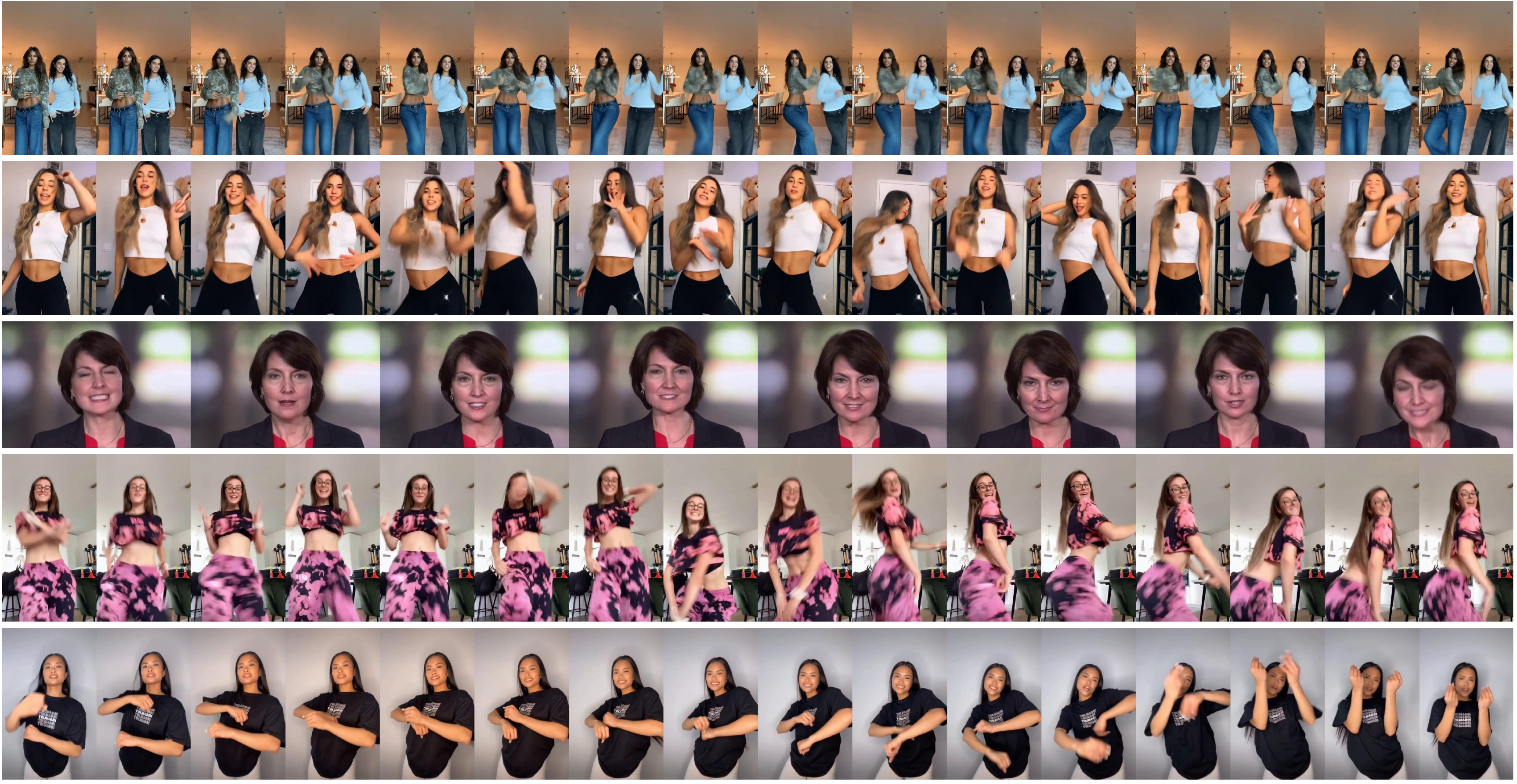}
	\caption{Some sample videos in our FakeHumanVid dataset.}
	\label{dataset_yangli}
    \vspace{-16pt}
\end{figure*}

\textbf{Real Video Collection:} There are already existing human-centric real video datasets~\cite{Jafarian_2021_CVPR_TikTok,zhang2021hdtf}. We aim to select real samples that are as similar as possible in content to fake videos. We observe that pose-driven generation methods mainly produce human dancing videos, audio-driven methods focus on generating talking face videos, and text-driven methods can create any open-ended video content. To match these with real videos, we collected raw videos from public datasets TikTok~\cite{Jafarian_2021_CVPR_TikTok} and HDTF~\cite{zhang2021hdtf}. These videos were then segmented into 5-10 second clips, finally constructing 7.6k real video clips.

\textbf{Fake Video Construction:} We design specific inference pipelines tailored to different video generation approaches, as illustrated in Figure~\ref{dataset}. For the pose-driven scenario, human poses extracted from real dance videos~\cite{Jafarian_2021_CVPR_TikTok} are adopted as the driving condition. In the audio-driven case, we separate audio tracks from real speaker videos~\cite{zhang2021hdtf} and utilize them as inputs. For text-driven synthesis, we first prompt Qwen2.5-VL~\cite{Qwen2.5-VL} to generate textual descriptions of real dance videos~\cite{Jafarian_2021_CVPR_TikTok}, which subsequently serve as driving conditions. Across all three scenarios, the first frame from each original video is used as the reference image. The duration of generated videos is restricted to 5-10 seconds, either by limiting the provided condition or applying explicit length constraints. Consequently, we produce a total of 7.6k fake video clips. Figure~\ref{dataset_yangli} presents several video samples from our FakeHumanVid dataset. 
\textit{More examples can be found in the Appendix~\ref{Appendix More Examples}.}

\subsection{Preliminary of Group Relative Policy Optimization}
\label{Preliminary of Group Relative Policy Optimization}

GRPO is an advanced reinforcement learning method derived from Proximal Policy Optimization (PPO). Unlike PPO, which utilizes a separate critic model to estimate value functions explicitly, GRPO directly compares groups of candidate responses, thus significantly reducing computational complexity and enhancing training efficiency. Given an input query $q$, GRPO samples $N$ candidate responses $\{o_1, o_2, \dots, o_N\}$ from the current policy $\pi_{\theta_{old}}$, evaluating their quality through rewards $\{r_1, r_2, \dots, r_N\}$ provided by a reward function. GRPO computes the relative quality or advantage of each response $\hat{A}_{i,t}$ by normalizing its reward using the group's mean and standard deviation. After obtaining $\hat{A}_{i,t}$, the optimization objective $\mathcal{J}_{\text{GRPO}}(\theta)$ aims to maximize the expected relative advantage while constraining policy deviations from a reference model $\pi_{ref}$ using Kullback-Leibler (KL) divergence regularization. Specifically, the objective is formulated as follows:
\vspace{-10pt}

\begin{align}
&\mathcal{J}_{\text{GRPO}}(\theta)
= \mathbb{E}_{[q \sim Q, \{o_i\}_{i=1}^N \sim \pi_{\theta_{\text{old}}}(o\|q)]} \notag \frac{1}{N}\sum_{i=1}^{N}\frac{1}{\lvert o_i\rvert}\sum_{t=1}^{\lvert o_i\rvert} \notag \\[2pt]
&\quad \Bigg\{\min\!\left[\frac{\pi_{\theta}^{i,t}}{\pi_{\theta_{\text{old}}}^{i,t}}\hat{A}_{i,t},
\, \operatorname{clip}\!\left(\frac{\pi_{\theta}^{i,t}}{\pi_{\theta_{\text{old}}}^{i,t}}, 1-\epsilon, 1+\epsilon\right)\hat{A}_{i,t}\right] \notag \\[2pt]
&\qquad\;\; - \beta \cdot \mathbb{D}_{\text{KL}}\!\big[\pi_{\theta}\,\|\,\pi_{\text{ref}}\big]\Bigg\}.
\end{align}

In this formulation, $\epsilon$ controls the clipping range to ensure stable policy updates, and $\beta$ is a regularization coefficient for KL-divergence to prevent excessive deviation from the reference policy.

\subsection{Overall Framework of AvatarShield}
\label{Overall Framework of AvatarShield}

\begin{figure*}[t!]
	\centering
	\includegraphics[width=1.0\linewidth]{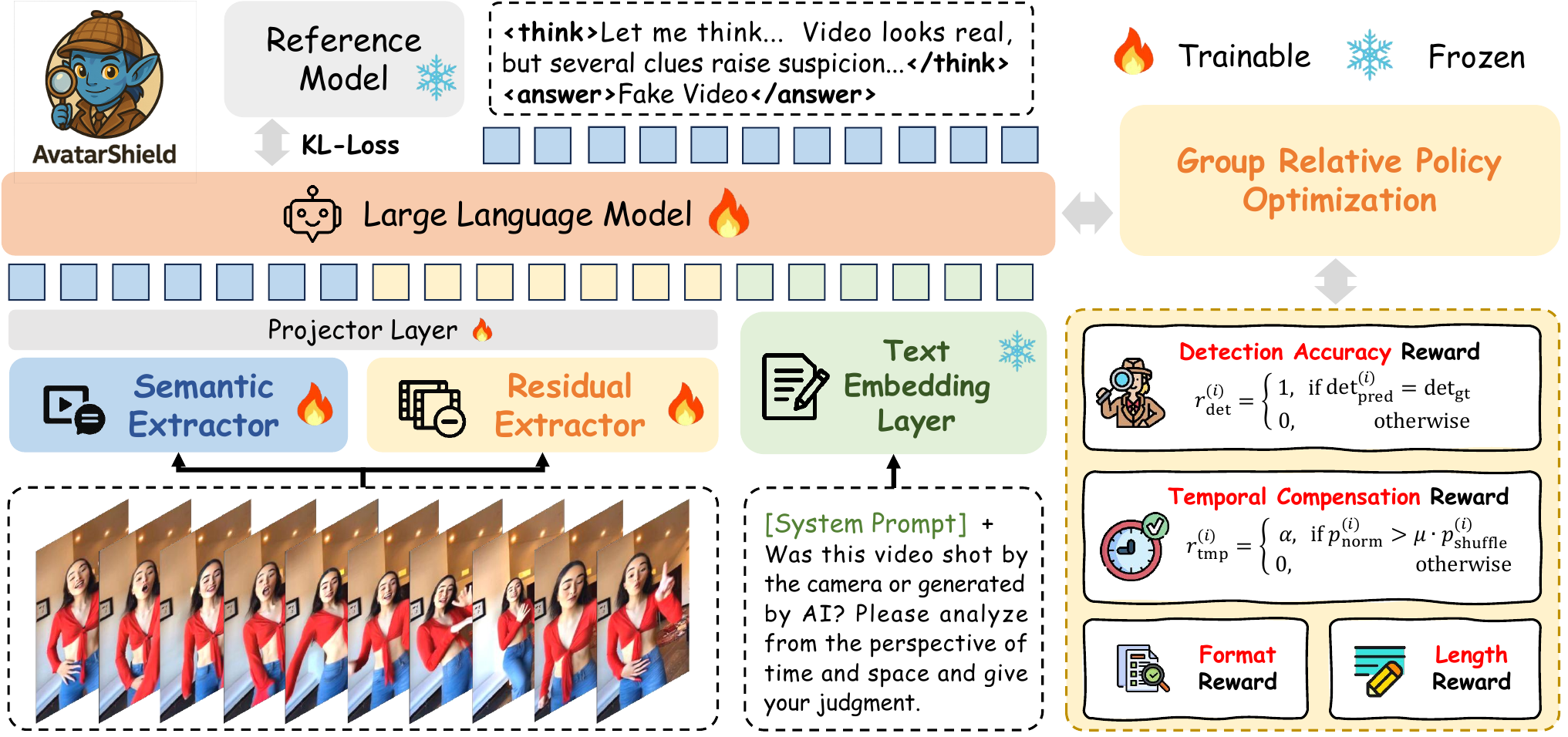}
	\vspace{-5pt}
	\caption{Illustration of the proposed AvatarShield. Our method takes text instructions as input through a text embedding layer and processes the video using a dual-encoder architecture, guiding the LLM to generate detection results along with reasoning outcomes. Then, under the GRPO framework, we jointly optimize the entire network through the detection accuracy reward, temporal compensation reward, format reward, and length reward, achieving precise and interpretable synthetic video detection.}
	\label{pipeline}
    \vspace{-16pt}
\end{figure*}

\textbf{Motivation:} Although current state-of-the-art general MLLMs~\cite{Qwen2.5-VL,chen2024internvl} have achieved remarkable performance on complex video understanding benchmarks, their design still presents limitations for AI-generated video detection. This is because their encoders are specifically built for high-level semantically related tasks such as action recognition or event reasoning, rather than for capturing the subtle low-level artifacts or distributional inconsistencies introduced by generative models. 
With the advancement of diffusion technologies, AI-generated videos have become increasingly realistic. Therefore, relying solely on semantic visual content may not be sufficient for accurately detecting videos produced by advanced generative methods~\cite{Klingai2024,brooks2024video}. Some researchers~\cite{wang2023dire} have found that real images reconstructed by a VAE show significantly more noticeable residual differences compared to images generated by diffusion models. (See Appendix~\ref{VAE Reconstruction Visualization} for details) Inspired by this observation, we propose a dual-encoder structure comprising a semantic extractor and a residual extractor. The semantic extractor captures high-level temporal dynamics, whereas the residual extractor identifies low-level local anomalies through residuals from VQ-VAE reconstruction. This dual-encoder architecture jointly models global semantic information and local detail artifacts, enabling more effective detection.

As illustrated in Figure~\ref{pipeline}, given an input video frame sequence $\mathbf{X} = \{x_t\}_{t=1}^T$, the framework employs a dual-encoder architecture for vision feature extraction and fusion. In the semantic extractor $\mathcal{E}_{\text{sem}}$, $\mathbf{X}$ is directly fed into a ViT to extract global semantic features, yielding the semantic feature sequence $\mathbf{F}_{\text{sem}}$. In the residual extractor $\mathcal{E}_{\text{res}}$, the original video frames undergo reconstruction via a VQ-VAE, producing the reconstructed video frames $\hat{\mathbf{X}}$. Subsequently, we compute the residual frames $\mathbf{R} = \lvert\mathbf{X} - \hat{\mathbf{X}}\rvert$ between the original and reconstructed frames. The residual sequence $\mathbf{R}$ is then input into a separate ViT to extract fine-grained residual features, resulting in the feature sequence $\mathbf{F}_{\text{res}}$.
Meanwhile, the system prompt $\mathbf{P}_{\text{sys}}$ (see Appendix~\ref{Prompts Design} for details) and user prompt $\mathbf{P}_{\text{user}}$ (e.g., "\textit{Was this video shot by the camera or generated by AI? Please analyze from the perspective of time and space and give your judgment.}") is processed by a text embedding layer $\mathcal{E}_{\text{text}}$, which encodes the instruction into a text feature sequence $\mathbf{F}_{\text{text}}$. This text embedding provides task-specific guidance to the model, helping steer the LLM toward interpretable and context-aware judgments.
Finally, the semantic feature $\mathbf{F}_{\text{sem}}$ and residual feature $\mathbf{F}_{\text{res}}$ are projected into multimodal tokens via dedicated projector layers and combined with the embedded text tokens $\mathbf{F}_{\text{text}}$. These fused representations are then fed into an LLM for reasoning and discriminating AI-generated content from authentic videos.
\begin{equation}
    \mathbf{F}_{\text{sem}} = \mathcal{E}_{\text{sem}}(\mathbf{X}), 
    \quad
    \mathbf{F}_{\text{text}} = \mathcal{E}_{\text{text}}(\mathbf{P}_{\text{sys}},\mathbf{P}_{\text{user}}), 
\end{equation}
\begin{equation}
    \mathbf{F}_{\text{res}} = \mathcal{E}_{\text{res}}(\lvert\mathbf{X}-\operatorname{VQ-VAE}(\mathbf{X})\rvert),
\end{equation}
\begin{equation}
    \mathbf{O}_{\text{det}} = \operatorname{LLM}(\phi_{\text{sem}}(\mathbf{F}_{\text{sem}}),\phi_{\text{res}}(\mathbf{F}_{\text{res}}),\mathbf{F}_{\text{text}}).
\end{equation}

Where $\phi_{\text{sem}}$ and $\phi_{\text{res}}$ respectively denote the sementic and residual projection layers. $\textbf{O}_{\text{det}}$ denotes the predicted reasoning and detection answers.

\subsection{Reward Functions for AIGC Fake Detection and Temporal Modeling}
\label{Reward Functions for AIGC Fake Detection and Temporal Modeling}


To effectively guide the learning process under the GRPO framework, we design a set of task-specific reward functions as shown in Figure~\ref{pipeline}. Our method includes four distinct rewards, namely detection accuracy reward, temporal compensation reward, length reward and format reward.




\textbf{Detection Accuracy Reward:} To support our human-centric fake detection task, we employed a detection reward based on the correctness of the predicted class label in comparison to the ground-truth. It serves as a straightforward but essential signal that directly aligns with the core detection objective. Suppose the detection result of the $i$-th response, denoted as \( \text{det}^{(i)}_{\text{pred}} \), matches the ground truth label $\text{det}_{\text{gt}}$; the reward is set to 1, otherwise it is set to 0. Specifically, the detection accuracy reward is formulated as:  
\begin{equation}
r^{(i)}_{\text{det}} = 1, \quad \text{if } \text{det}^{(i)}_{\text{pred}} = \text{det}_{\text{gt}}~\text{ else } 0.
\end{equation}




\textbf{Temporal Compensation Reward:} One of the major and challenging issues in current AI-generated video content is the difficulty in maintaining temporal consistency across frames. While many advanced video generation models are capable of producing visually impressive individual frames, they often struggle with preserving smooth inter-frame coherence. This results in issues such as flickering, unnatural transitions, and abrupt changes in object presence or motion. These temporal artifacts are particularly problematic in human-centric synthetic videos, where even subtle inconsistencies in motion patterns, timing, or object interactions can significantly detract from the overall realism.
Motivated by this observation, we propose a temporal compensation reward to further enhance the LLM's ability to capture temporal cues and model motion patterns effectively. Specifically, given the same question $q$, we input the original and residual videos with normally ordered tokens into the LLM, obtaining a group of answers \(o_{\text{norm}}\). Simultaneously, we randomly shuffle the two processed video tokens and also feed them into the model, resulting in another group of answers \(o_{\text{shuffle}}\). If the probability of all answers being correct for the ordered sequence \(p_{\text{norm}}\) is greater than that of the shuffled sequence \(p_{\text{shuffle}}\), we infer that the model successfully captures temporal relationships and motion patterns within the video, thus granting it a temporal compensation reward \(r^{(i)}_{\text{tmp}}\). The temporal compensation reward is defined as:

\begin{equation}
r^{(i)}_{\text{tmp}} = \alpha, \quad \text{if } p^{(i)}_{\text{norm}} > \mu \cdot p^{(i)}_{\text{shuffle}}~\text{ else } 0.
\end{equation}

Where \(\alpha\), \(\mu\) are respectively set to $0.3$ and $0.8$, which encourages the model to rely on temporal reasoning. It explicitly strengthens the model's ability to leverage temporal information for accurate fake detection by comparing its performance on ordered and shuffled sequences. 

\textbf{Length Reward:} To ensure that our model strikes a balance between deep reasoning and overthinking, we introduce a length reward mechanism. Given a reasoning path \(o^{(i)}\), if its output length falls within the range \([l_{\text{min}}, l_{\text{max}}]\), the model is granted an additional reward. Formally, the reward is expressed as:
\begin{equation}
    r^{(i)}_{\text{len}} = \lambda, \quad \text{if } l_{\text{min}} \leq \operatorname{length}(o^{(i)}) \leq l_{\text{max}}~\text{else } 0.
\end{equation}

Where \(\lambda\) is set to $0.1$ used to encourage effective and concise reasoning. $l_{\text{min}}$, $l_{\text{max}}$ are respectively set to $320$ and $512$.

\textbf{Format Reward:} Format reward is designed to ensure that the model's responses follow a well-structured format. Specifically, the model is required to present its reasoning paths enclosed within ``<think>...</think>'' tags, and its final prediction within ``<answer>...</answer>'' tags. The reward $r^{(i)}_{\text{fmt}}$ is set to $1$ if the $i$-th response $o^{(i)}$ fulfills all the above conditions; otherwise, its reward is $0$.

Finally, we perform a linear combination of these four rewards to jointly guide the optimization of AvatarShield, inspiring it to discover traces of forgery in videos via deep reasoning.

\begin{table*}[t!]
\centering
\caption{Comparison results of synthetic video detection on in-domain data between our method and competitive methods. Our method outperforms all other methods across various pose-driven, audio-driven, and text-driven generation videos. [S.A.: StableAnimator, M.M.: MimicMotion, C.N.: ControlNeXt]}
\vspace{5pt}
\label{table:in_domain}
\renewcommand{\arraystretch}{1.5}
\resizebox{1.\linewidth}{!}{
\begin{tabular}{c|c|ccccccc}
\shline
Category & Method & CNNSpot~\cite{wang2019cnnspot} & VSwinT~\cite{liu2021video} & DIRE~\cite{wang2023dire} & Uni-FD~\cite{ojha2023fakedetect} & HiFi-Net~\cite{guo2023hierarchical} & MM-Det~\cite{song2024learning} & Ours \\ 
\hline
\multirow{3}{*}{Pose-Driven}    & S.A.~\cite{tu2024stableanimator} & \secondbest{0.8994} & 0.8586   & 0.8638  & 0.8767  & 0.8702  & 0.8903  & \best{0.9573 } \\ 
                                & M.M.~\cite{zhang2024mimicmotion} & 0.8256 & 0.8718   & 0.8608  & 0.8275  & 0.8788  & \secondbest{0.9260}   & \best{0.9483} \\ 
                                & C.N.~\cite{peng2024controlnext} & 0.7821 & 0.8460   & 0.7609  & 0.8356  & 0.7198  & \secondbest{0.9258}  & \best{0.9333 } \\ 
\hline
\multirow{2}{*}{Audio-Driven}   & Hallo3~\cite{cui2024hallo3} & 0.7249 & \secondbest{0.8060}   & 0.7466  & 0.7790 & 0.6455  & 0.7671 & \best{0.8897 } \\ 
                                & HelloMeme~\cite{zhang2024hellomeme} & \secondbest{0.9331 } & 0.8204  & 0.9098  & 0.8521  & 0.7130  & 0.9100  &  \best{0.9808 } \\ 
\hline
\multirow{4}{*}{Text-Driven}    & Kling~\cite{Klingai2024} & 0.8482 & 0.8110   & 0.7843  & 0.8008  & 0.7646  & \secondbest{0.8913 }  & \best{0.9192 } \\ 
                                & Hailuo~\cite{hailuo} & 0.8001 & 0.7306   & 0.7552 & 0.7554 & 0.7663 & \secondbest{0.8750} & \best{0.8865} \\ 
                                & Wanx~\cite{wang2025wan}  & 0.8824 & 0.9039  & 0.7992 & 0.9690 & 0.7924 & \secondbest{0.9563}  & \best{0.9706} \\ 
                                & CogVideo~\cite{yang2024cogvideox} & 0.8502 & 0.8788  & 0.8914 & 0.8388 & 0.8359 & \secondbest{0.9396} & \best{0.9571} \\ 
\hline
\multicolumn{2}{c|}{Mean}       & 0.8384 & 0.8363 & 0.8191 & 0.8372 & 0.7763 & \secondbest{0.8979} & \best{0.9381}  \\
\shline
\end{tabular}}
\vspace{-10pt}
\end{table*}

\section{Experiments}
\label{Experiments}

\subsection{Experimental Setup}
\label{Experimental Setup}

\textbf{Dataset:} In our experiments, we use the proposed FakeHumanVid benchmark for both training and evaluation. As shown in Figure~\ref{dataset}, the benchmark consists of videos generated by different generators along with their corresponding real videos. For fairness, we mix each generated video with its corresponding real video that provided the generation condition, ensuring a 1:1 ratio of real to fake videos in every dataset. Furthermore, each dataset is split into training and testing subsets following a 9:1 ratio. The datasets are referred to by the name of the generator, and unless otherwise specified, we do not report performance on real videos separately. \textit{More details are provided in the Appendix~\ref{Appendix Dataset Construction}}.

\textbf{State-of-the-Art Methods:} To ensure a fair comparison, we selected mainstream baseline methods that provide open-source code or pretrained models, including CNNSpot~\cite{wang2019cnnspot}, the first work to use ResNet for AI-generated image detection; VSwinT~\cite{liu2021video}, which uses the transformer's global attention mechanism to capture spatial and temporal video information; DIRE~\cite{wang2023dire}, which introduces the use of DDIM-based~\cite{song2021denoising} reconstruction for efficient diffusion image detection; Uni-FD~\cite{ojha2023fakedetect}, which employs a CLIP-like~\cite{radford2021learning} architecture for effective fake image detection; HiFi-Net~\cite{guo2023hierarchical}, which uses a multi-branch feature extraction module to enhance the detection of synthetic images; and MM-Det~\cite{song2024learning}, which incorporates a dynamic fusion strategy to leverage the forgery representation capabilities of MLLMs effectively.

\textbf{Implementation Details:} We initialize our model with Qwen2.5-VL-7B~\cite{Qwen2.5-VL} and perform full-parameter fine-tuning using the R1-V framework~\cite{chen2025r1v}. The model is trained for one epoch on $8$ NVIDIA A800 80G GPUs, with a learning rate of $1$$\times$$10^{-6}$. For the GRPO training setting, we select $\beta=0.04$ to constrain the policy model and the reference model. For the evaluation metrics, we employ AUC to measure the detection accuracy of the methods.

\begin{figure*}[t!]
	\centering
	\includegraphics[width=1.0\linewidth]{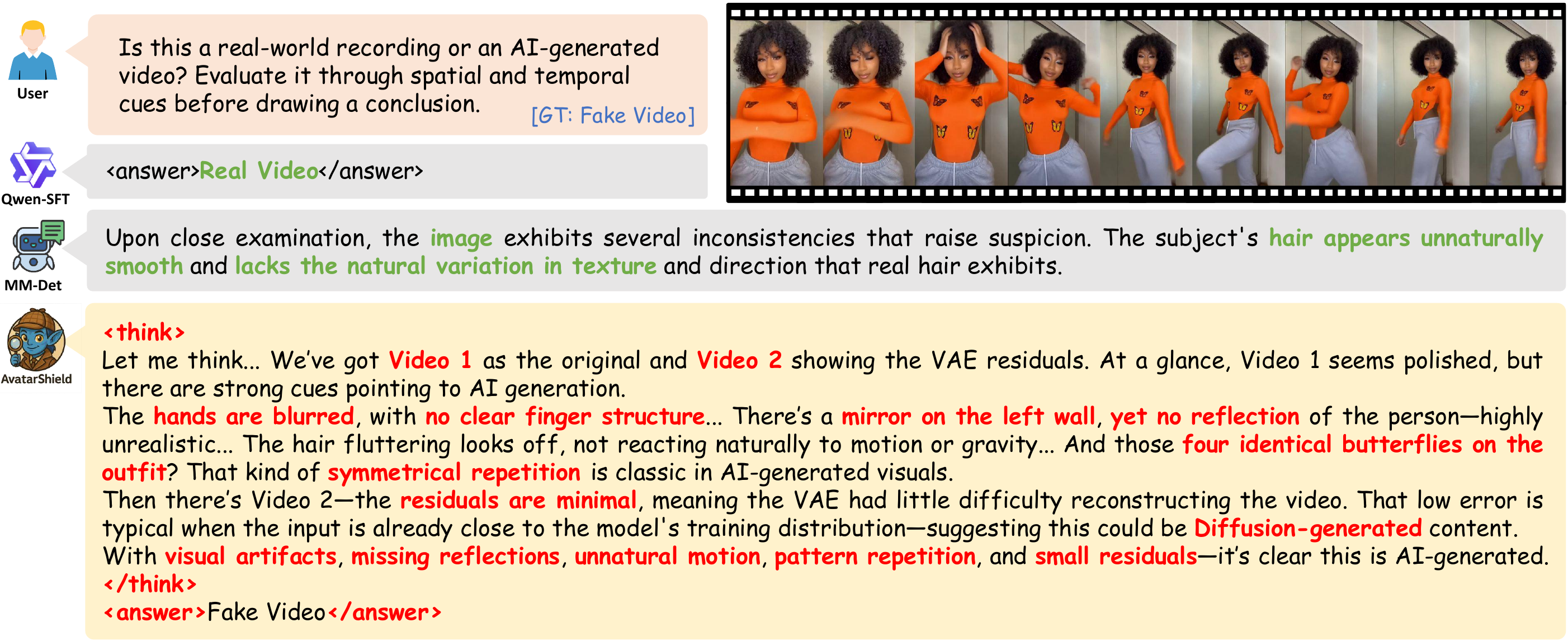}
	\vspace{-10pt}
	\caption{Comparison results between our method and other LLM-based methods. While Qwen-SFT can only output binary real-or-fake judgments, and MM-Det can only provide fake information analysis for each frame, our method not only delivers accurate detection results but also provides a detailed and transparent reasoning process.}
	\label{zhuguan}
    \vspace{-16pt}
\end{figure*}

\subsection{In-Domain Detection Results}
\label{In-Domain Detection Results}

To evaluate the detection accuracy of our detection method on the in-domain data, we trained our AvatarShield and other comparison methods on all 9 training sets and tested on the respective testing sets. The results are shown in Table~\ref{table:in_domain}. We observe that our method achieves state-of-the-art performance across all test sets, consistently outperforms all competitive approaches.
Notably, AvatarShield demonstrates a large performance margin in several cases. On the HelloMeme dataset, our method achieves an AUC of 0.9808, significantly outperforming the second-best method CNNSpot. Similarly, for the pose-driven method StableAnimator, AvatarShield reaches 0.9573, surpassing the next-best result by 5.79\%. Another substantial lead is observed on the text-driven method Kling, where our method achieves 0.9192, compared to the second-best score of 0.8913 by MM-Det. 

Additionally, Figure~\ref{zhuguan} displays the responses from several LLM-based methods. Although MM-Det is an LLM-based method, it processes videos frame-by-frame using a MLLM without considering temporal information. As a result, its responses are based solely on individual frames and are often simple and vague, such as ``hair appears unnaturally smooth''. In contrast, when SFT is applied to fine-tune Qwen~\cite{Qwen2.5-VL}, the model provides an answer directly, but the correctness is not guaranteed. Our method, however, first combines both the spatial and temporal information of the video, offering a detailed and specific reasoning process. It can detect issues such as ``visual artifacts, missing reflections, unnatural motion,'' and ultimately provides an accurate answer. The superior performance of AvatarShield arises from our dual-encoder architecture with capabilities to capture complex spatio-temporal features, and our efficient reward functions design. \textit{More comparison results and dialog examples are detailed in the Appendix~\ref{Appendix More Examples}.}

\subsection{Cross-Domain Detection Results}
\label{Cross-Domain Detection Results}

\begin{table*}[t!]
\centering
\caption{Comparison results of synthetic video detection on cross-domain data between our method and competitive methods. [H.M.: HelloMeme]}
\label{table:cross_domain}
\renewcommand{\arraystretch}{1.5}
\resizebox{1.\linewidth}{!}{
\begin{tabular}{c|c|ccccccccc|c}
\shline
\multicolumn{2}{c}{Dataset} & S.A.~\cite{tu2024stableanimator} & M.M.~\cite{zhang2024mimicmotion} & C.N.~\cite{peng2024controlnext} & Hallo3~\cite{cui2024hallo3} & H.M.~\cite{zhang2024hellomeme} & Kling~\cite{Klingai2024} & Hailuo~\cite{hailuo} & Wanx~\cite{wang2025wan} & CogVideo~\cite{yang2024cogvideox} & Mean \\ 
\hline
\multirow{3}{*}{DIRE} & Pose & 0.7528 & 0.8230 & 0.8276 & 0.6663 & 0.7737 & 0.6655 & 0.7286 & 0.7443 & 0.7472 & 0.7477  \\ 
& Text & 0.7904 & 0.7985 & 0.7331 & 0.6672 & 0.6499 & 0.7936 & 0.8362 & 0.7646 & 0.8301 & 0.7626  \\ 
& Mix  & 0.7677 & 0.7895 & 0.7694 & 0.7987 & 0.8019 & 0.7688 & 0.8524 & 0.7526 & 0.7835 & 0.7872  \\ \hline
\multirow{3}{*}{MM-Det} & Pose & 0.8248 & 0.8627 & 0.8308 & 0.7505 & 0.8346 & 0.7192 & 0.7577 & 0.6483 & 0.8143 & \secondbest{0.7825}  \\ 
& Text & 0.6926 & 0.7285 & 0.7778 & 0.6766 & 0.7122 & 0.8385 & 0.8429 & 0.7999 & 0.8429 & \secondbest{0.7680}  \\ 
& Mix  & 0.8217 & 0.8690 & 0.8111 & 0.7881 & 0.7646 & 0.7885 & 0.8610 & 0.7845 & 0.8143 & \secondbest{0.8114}  \\ \hline
\multirow{3}{*}{Ours}  & Pose & 0.9023 & 0.9351 & 0.8444 & 0.8592 & 0.8962 & 0.8500 & 0.8393 & 0.8391 & 0.8571 & \best{0.8692}  \\ 
& Text & 0.8782 & 0.8688 & 0.8111 & 0.8576 & 0.8038 & 0.9000 & 0.8834 & 0.9134 & 0.8857 & \best{0.8669}  \\ 
& Mix  & 0.8978 & 0.9398 & 0.8556 & 0.8845 & 0.8808 & 0.8308 & 0.9036 & 0.8832 & 0.8286 & \best{0.8783}  \\ 
\shline
\end{tabular}}
\vspace{-10pt}
\end{table*}

\begin{table}[t!]
\centering
\renewcommand{\arraystretch}{1.5}
\caption{Performance comparison with DeepFake detection methods.}
\begin{tabular}{c|c|c|c|c}
\shline
Methods & CADDM & Exposing & RECCE & Ours \\
\hline
Hallo3~\cite{cui2024hallo3} & 0.7767 & 0.7920 & \secondbest{0.8125} & \best{0.8897} \\
H.M.~\cite{zhang2024hellomeme} & 0.8421 & \secondbest{0.9103} & 0.9077 & \best{0.9808} \\
\shline
Mean & 0.8094 & 0.8512 & \secondbest{0.8601} & \best{0.9353} \\
\shline
\end{tabular}
\label{tab:comparison_dffd_seqdeepfake_transposed}
\end{table}

To evaluate the generalization ability of our method on unseen videos, we design a set of cross-domain comparison experiments to simulate realistic scenarios. We select MM-Det~\cite{song2024learning} and DIRE~\cite{wang2023dire} as comparison methods, which are respectively represent the LLM-based and non-LLM-based synthetic detectors.
All three methods are trained on only one subset of FakeHumanVid: pose-driven (StableAnimator~\cite{tu2024stableanimator}, MimicMotion~\cite{zhang2024mimicmotion}, ControlNeXt~\cite{peng2024controlnext}), text-driven (Kling~\cite{Klingai2024}, Hailuo~\cite{hailuo}, Wanx~\cite{wang2025wan}, CogVideo~\cite{yang2024cogvideox}), or a mixed group that combines one method from each category (StableAnimator~\cite{tu2024stableanimator}, Hallo3~\cite{cui2024hallo3}, Hailuo~\cite{hailuo}), and are evaluated across all $9$ generation methods. This setup allows us to comprehensively assess the model's generalization and zero-shot detection performance in a realistic application scenario. The mixed group setting is particularly meaningful, as most current forgery methods are developed along the pose-driven, text-driven, or audio-driven paradigms, and in practice the available training data may only cover one representative generation method from each domain. By training on such a mixed subset, we can better evaluate the detector’s ability to generalize to other unseen generation methods within all domains, providing a strong indication of its robustness in real-world deployments.
Note that, due to the limited scale of audio-driven data, it is challenging to effectively fine-tune the LLM. We plan to expand the dataset in future work to address this issue. 

As reported in Table~\ref{table:cross_domain}, AvatarShield consistently outperforms both MM-Det and DIRE across all evaluation settings, including pose-driven, text-driven, and mixed training scenarios. When trained on pose-driven data, AvatarShield shows substantial improvements over the two comparison methods, with particularly large gains on challenging datasets such as Hallo3 and Wanx. For example, its AUC on Hallo3 rises from 0.7505 in MM-Det and 0.6663 in DIRE to 0.8592, while on Wanx it improves from 0.6483 in MM-Det and 0.7443 in DIRE to 0.8391. Under text-driven training, AvatarShield maintains strong generalization ability, achieving higher performance than the comparison methods on nearly all test sets. Notably, its AUC on Wanx increases to 0.9134 compared with 0.7999 in MM-Det and 0.7646 in DIRE, and on CogVideo it improves to 0.8857, exceeding 0.8429 in MM-Det and 0.8301 in DIRE. When trained on the mixed subset, AvatarShield achieves the highest AUC values in almost all cases, such as 0.9398 on MimicMotion and 0.9036 on Hailuo, which are considerably higher than 0.8690 and 0.8610 in MM-Det and 0.7895 and 0.8524 in DIRE. The Mean column further confirms AvatarShield’s consistent advantage, showing the highest overall average performance across all training settings. These results demonstrate that AvatarShield not only surpasses the strong LLM-based detector MM-Det but also significantly outperforms the classical non-LLM-based method DIRE, highlighting its superior generalization and zero-shot detection capability. 
This performance gain can be attributed to AvatarShield’s GRPO optimization, which dynamically adapts detection strategies via reinforcement learning, and our designed temporal compensation reward, which guides the LLM to perceive inter-frame consistency in videos, making it robust and reliable in real-world scenarios.

\begin{table*}[t!]
\centering
\renewcommand{\arraystretch}{1.5}
\caption{Ablation Studies on the key components of our AvatarShield, where ``res'' denotes residual videos, ``recons'' denotes reconstructed videos, and ``TCR'' denotes temporal compensation reward.}
\label{table:ablation}
\begin{adjustbox}{max width=\textwidth}
\begin{tabular}{l|ccccccccc|cc}
\shline
Method & S.A.~\cite{tu2024stableanimator} & M.M.~\cite{zhang2024mimicmotion} & C.N.~\cite{peng2024controlnext} & Hallo3~\cite{cui2024hallo3} & H.M.~\cite{zhang2024hellomeme} & Kling~\cite{Klingai2024} & Hailuo~\cite{hailuo} & Wanx~\cite{wang2025wan} & CogVideo~\cite{yang2024cogvideox} & Mean \\ 
\hline
w/o res & 0.7176  & 0.8604 & 0.8586 & 0.7984 & 0.8218  & 0.8019 & 0.8159 & 0.7640  & 0.7518 & 0.7989  \\ 
w/ recons  & 0.8204  & 0.8744 & 0.7201 & 0.8739 & 0.9223  & 0.8848 & 0.8651 & 0.9070  & 0.8797 & 0.8609  \\ 
w/o TCR & 0.7522  & 0.8870 & 0.8405 & 0.8286 & 0.9236  & 0.8213 & 0.8166 & 0.9147  & 0.8210 & 0.8451   \\
SFT & 0.8203 & 0.8437 & 0.7945 & 0.7472 & 0.9485 & 0.7913 & 0.8008 & 0.8786 & 0.8460 & 0.8301  \\ \hline
Ours & 0.9573  & 0.9483 & 0.9333 & 0.8897 & 0.9808  & 0.9192 & 0.8865 & 0.9706 & 0.9571 & 0.9381   \\ 
\shline
\end{tabular}
\end{adjustbox}
\vspace{-10pt}
\end{table*}

\subsection{Comparison with DeepFake Detection Methods}


Considering that the audio-driven data in the FakeHumanVid benchmark partially overlaps with the facial reenactment techniques~\cite{prajwal2020lip} considered in DeepFake detection methods, as both utilize speech audio to drive a static face image to generate talking-head videos, we conduct a comparative evaluation of our method against existing DeepFake detection approaches under in-domain setting. As shown in Table~\ref{tab:comparison_dffd_seqdeepfake_transposed}, we benchmark our method against three representative DeepFake detection baselines: CADDM~\cite{dong2023implicit}, Exposing~\cite{ba2024exposing}, and RECCE~\cite{cao2022end}, across audio-driven datasets: Hallo3 and HelloMeme (H.M.).

From the results, we observe that our proposed method significantly outperforms all competing methods on both datasets. Notably, while RECCE achieves the strongest performance among baselines (0.8125 on Hallo3 and 0.9077 on H.M.), our method surpasses it with large margins, reaching 0.8897 and 0.9808 respectively. These results highlight our model's superior capability in capturing subtle temporal and spatial inconsistencies commonly present in AI-generated facial motion, making it more effective at distinguishing real from synthetic videos in audio-driven settings.

\subsection{Ablation Study}
\label{Ablation Study}

To evaluate the effect of our residual extractor, temporal compensation reward, and GRPO strategy in enhancing the LLM's ability to detect AI-generated videos, we conduct ablation studies with four variants. First, we remove the residual extractor, allowing the LLM to receive input solely from the semantic extractor. In the second, we disable the residual computation within the residual extractor, so the LLM receives the VAE-reconstructed video directly. In the third variant, we remove the temporal compensation reward to assess its individual contribution. In the fourth variant, we omit the GRPO algorithm and directly fine-tune the framework using the SFT method. All models are trained under the same in-domain detection settings as AvatarShield.

The results are summarized in Table~\ref{table:ablation}. We observe that each component contributes meaningfully to the overall performance of our method. Removing the residual extractor causes a significant drop in AUC across all datasets, indicating that direct extraction of generation-specific artifacts is crucial for effective detection. When the model uses only VAE-reconstructed videos without computing residuals, performance improves slightly but remains clearly lower than the full model, suggesting that residual signals are more discriminative than raw reconstructions. Eliminating the temporal compensation reward also leads to consistent performance degradation, particularly on videos with temporal inconsistencies such as StableAnimator and CogVideo. Moreover, using SFT for fine-tuning the model results in a marked performance drop compared to our approach, highlighting the essential role of GRPO in enhancing the model’s ability to generalize across various video generation methods. These results demonstrate that the semantic extractor, residual extractor, and temporal reward mechanism work together to enhance the model’s ability to detect both spatial and temporal artifacts in AI-generated videos, yielding the best results when all components are combined.

\section{Conclusion}
\label{Conclusion}

In this paper, we propose \textbf{AvatarShield}, a novel multimodal human-centric synthetic video detection framework that integrates GRPO into the training of LLM. To the best of our knowledge, this is the first attempt to introduce GRPO-based reinforcement learning into AI-generated video detection. By leveraging this innovative training paradigm, we successfully stimulate and enhance the intrinsic reasoning and perceptual abilities of LLM, thereby reducing reliance on costly supervised fine-tuning.
Our proposed method achieves state-of-the-art performance across both in-domain and cross-domain evaluation settings, significantly outperforming existing baselines. Notably, the cross-domain experiments, designed to simulate real-world scenarios involving unseen forgery methods, demonstrate the robust generalization capability of AvatarShield. This highlights the practical applicability and resilience of our system in addressing the evolving threats posed by AI-generated video content.
Looking ahead, AvatarShield has the potential to contribute broadly across various application domains. These include but are not limited to digital media forensics, social media content moderation, legal and law enforcement investigations, and the protection of personal portrait rights. By providing a scalable and generalizable detection solution, our work lays a solid foundation for building trustworthy AI systems capable of safeguarding visual content authenticity in an increasingly synthetic media landscape.

\textbf{Limitations: }Despite the strong performance of our method, it still has certain limitations. The inference speed of large language models on long video sequences remains relatively slow, which can be solved by using LLM acceleration and pruning methods~\cite{kwon2023efficient}. Additionally, although our method shows good generalization across existing generation methods, its robustness against future techniques with more subtle artifacts still needs further exploration.
\vspace{-10pt}


\begin{appendices}

\section{Dataset Construction}
\label{Appendix Dataset Construction}

Our FakeHumanVid dataset is constructed using 9 distinct video generation methods, covering three major modalities: text-driven, pose-driven, and audio-driven video synthesis. Among them, we employed the official APIs of four production-grade models, namely Kling I2V version 1.6~\cite{Klingai2024}, Hailuo I2V-01 live~\cite{hailuo}, Wanx I2V version 2.1~\cite{wang2025wan}, and CogVideo-X2~\cite{yang2024cogvideox}, to ensure stable access to their latest functionalities. The remaining five methods, including StableAnimator~\cite{tu2024stableanimator}, MimicMotion~\cite{zhang2024mimicmotion}, ControlNeXt~\cite{peng2024controlnext}, Hallo 3~\cite{cui2024hallo3}, and HelloMeme~\cite{zhang2024hellomeme}, were reproduced locally using their official open-source code and pretrained weights available on GitHub. This hybrid strategy balances quality and reproducibility, providing a rich variety of video styles. The construction process for each method is described in detail below.

\textbf{Kling (I2V v1.6)}~\cite{Klingai2024}: Kling 1.6, developed by Kuaishou Technology, is an advanced image-to-video generation model. It transforms static images into dynamic 5-second videos at 720p resolution, offering high-quality visual outputs with enhanced motion and semantic understanding. This version introduces significant improvements over its predecessor, Kling 1.5, making it a standout tool for content creators looking to transform static images into dynamic video content.

\textbf{Hailuo (I2V-01-live)}~\cite{hailuo}: Hailuo I2V-01 Live is a specialized image-to-video (I2V) model that converts still images into animated video sequences. It maintains consistency across frames while providing smooth motion and precise control over facial expressions and camera movements. This model is specifically trained for Live2D and general animation use cases.

\textbf{Wanx (I2V v2.1)}~\cite{wang2025wan}: Wanx 2.1 is an open-source AI video generation model based on Diffusion Transformer and Wan-VAE. It supports various tasks like text-to-video (T2V), image-to-video (I2V), and more. Wanx 2.1 offers superior performance, multi-tasking capabilities, and consumer-grade GPU compatibility, setting a new standard for video generation.

\textbf{CogVideo-X2}~\cite{yang2024cogvideox}:  CogVideo-X2 is a text-to-video generation model focused on creating more coherent videos aligned with a prompt. It achieves this using several methods, including a 3D variational autoencoder that compresses videos spatially and temporally, improving compression rate and video accuracy.

\textbf{StableAnimator}~\cite{tu2024stableanimator}: StableAnimator is a high-quality identity-preserving human image animation tool. It generates high-fidelity video based on reference images and pose sequences without post-processing. The model begins by computing image and face embeddings with off-the-shelf extractors and introduces a novel distribution-aware ID Adapter to prevent interference caused by temporal layers while preserving identity via alignment.

\textbf{MimicMotion}~\cite{zhang2024mimicmotion}: MimicMotion is a high-quality human motion video generation model with confidence-aware pose guidance. Developed by Tencent and Shanghai Jiao Tong University, it can generate detailed and realistic human motion videos from a single pose sequence image, handling various activities like dance, sports, or everyday actions effortlessly.

\textbf{ControlNeXt}~\cite{peng2024controlnext}: ControlNeXt is a controllable video and image generation model that supports various base models (SD1.5, SDXL, SD3, SVD) and tasks (image/video generation with various conditions). It introduces a lightweight controllable module that reduces trainable parameters by up to 90\% compared with ControlNet, achieving faster convergence and outstanding efficiency.

\textbf{Hallo 3}~\cite{cui2024hallo3}: Hallo 3 is an open-source portrait animation model developed by Fudan Vision Lab. It uses Diffusion Transformer Networks to generate realistic talking head videos from photos and audio. The model addresses challenges in handling non-frontal perspectives, rendering dynamic objects around the portrait, and generating immersive, realistic backgrounds.

\textbf{HelloMeme}~\cite{zhang2024hellomeme}: HelloMeme is an open-source portrait animation model that integrates spatial weaving attention mechanisms to provide high-quality image and video generation. It supports Gradio and ComfyUI interfaces, enabling a wide range of experiments and applications. HelloMeme is designed to generate localized high-fidelity expression-action-consistent images or videos. 

\begin{figure*}[t!]
	\centering
	\includegraphics[width=0.6\linewidth]{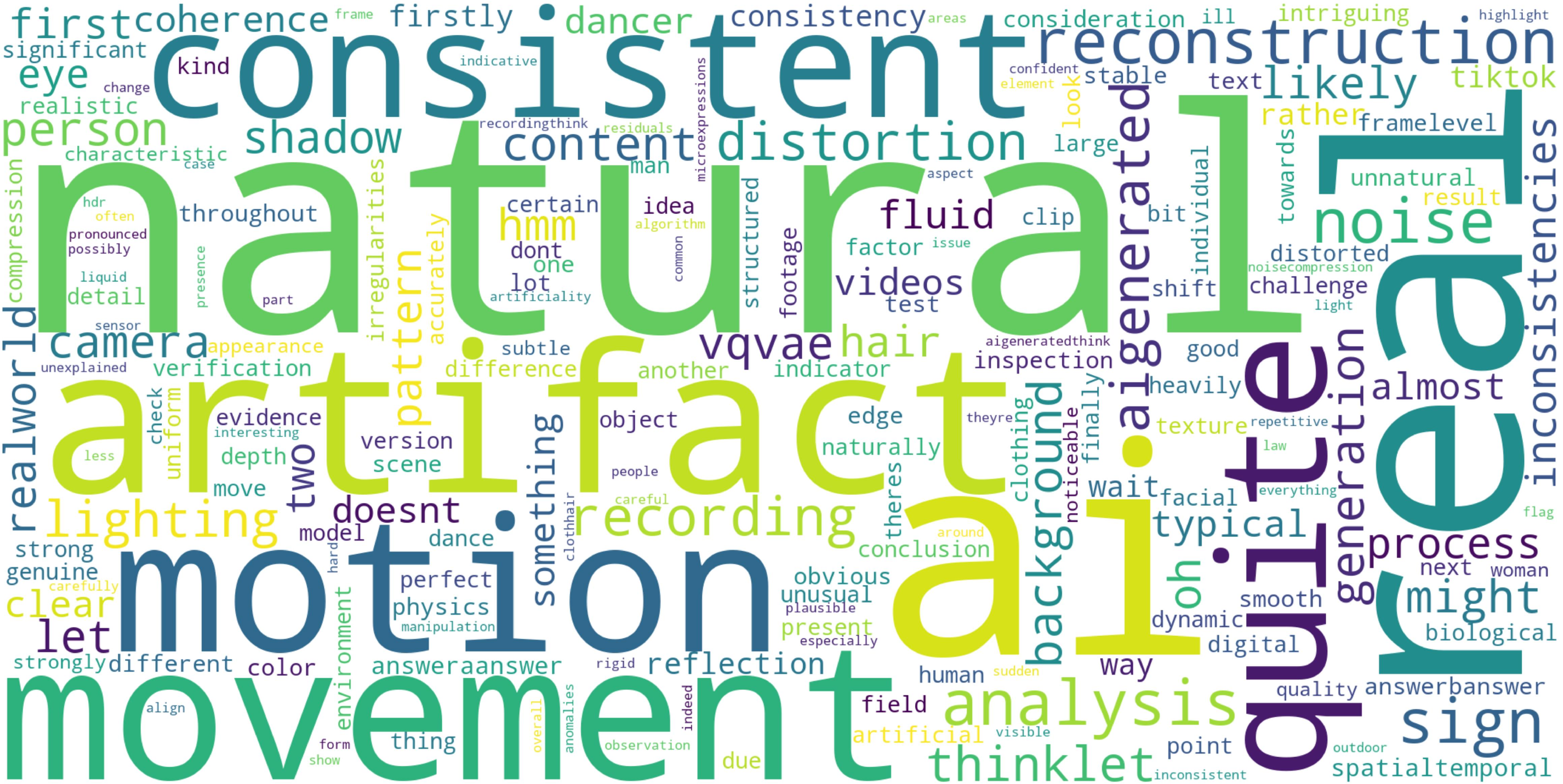}
	\caption{The nouns, adjectives and verbs word clouds of our AvatarShield. Our method can correctly output some analysis about motion, artifacts and authenticity of a suspected video.}
    \vspace{-15pt}
    \label{fig:ciyun}
\end{figure*}

\section{Answer Analysis}

To better understand the interpretive focus of the model during the detection of AI-generated videos, we conduct a linguistic analysis of the most frequent nouns, adjectives, and verbs in its explanations. The resulting word cloud, illustrated in Figure~\ref{fig:ciyun}, highlights dominant concepts and linguistic signals emphasized by the model. Key terms such as `natural', `artifact', `motion', `real', `consistent', and `movement' suggest that the model prioritizes spatial-temporal coherence and naturalistic dynamics when evaluating authenticity. Additionally, the recurrence of words like `background', `shadow', `lighting', and `reconstruction' reflects the model’s attention to low-level visual fidelity and physical plausibility. This analysis demonstrates that the model leverages comprehensive, interpretable visual patterns rather than superficial features, grounding its assessments in both structural and temporal inconsistencies commonly found in manipulated media.

\vspace{-10pt}

\section{Prompts Design}
\label{Prompts Design}

\begin{figure*}[t!]
	\centering
	\includegraphics[width=0.9\linewidth]{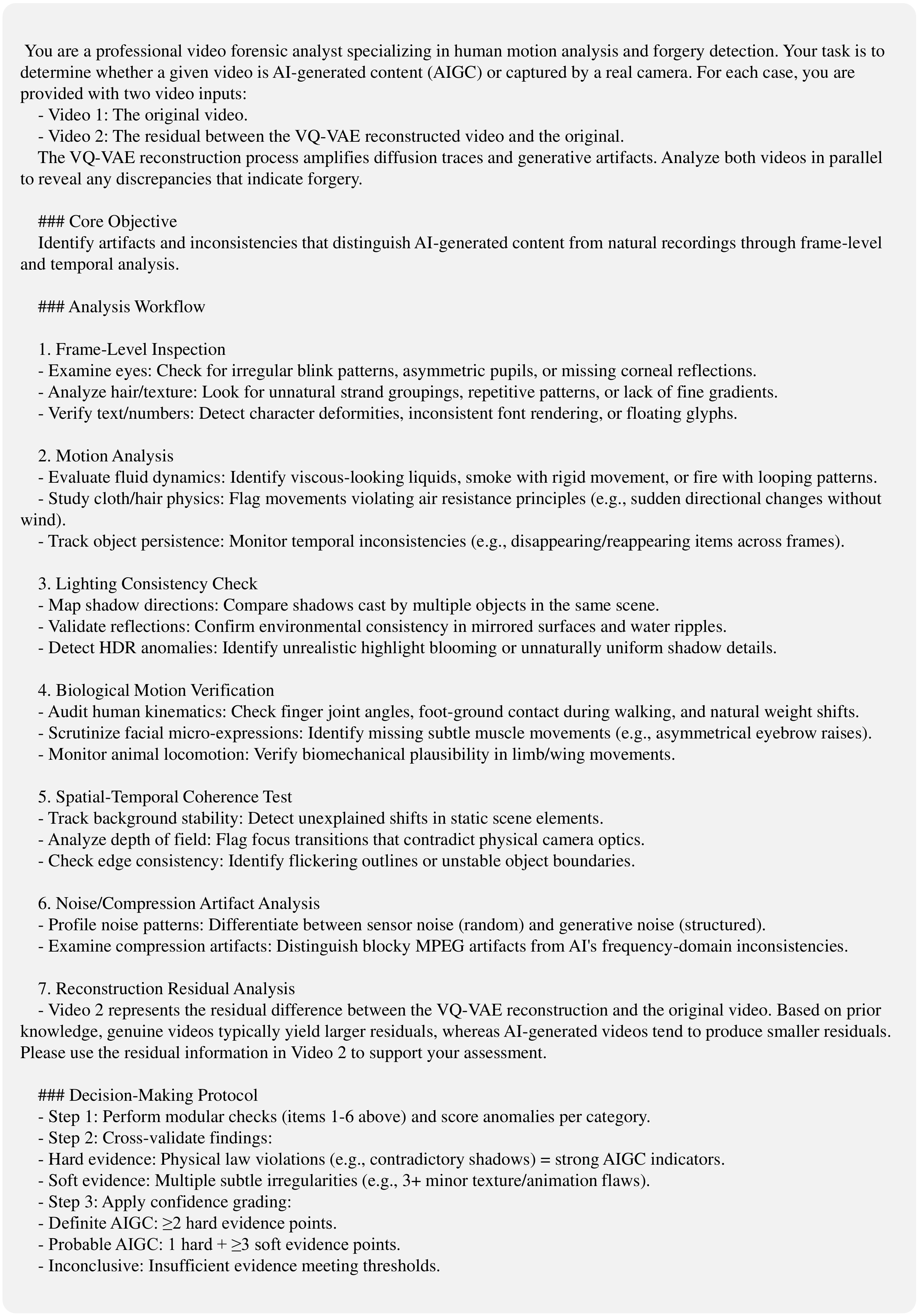}
	\caption{Illustrations of our detection system prompt design. We guide the LLM to comprehensively evaluate the authenticity of a video by considering aspects such as frame-level inspection, motion, and lighting.}
	\label{img_prompt}
    \vspace{-16pt}
\end{figure*}

During the training of AvatarShield, we carefully designed a dedicated system prompt to guide the MLLM in effectively determining whether a video is AI-generated, as illustrated in Figure~\ref{img_prompt}. This prompt strategically instructs the model to conduct a comprehensive analysis from seven key perspectives, including Frame-Level Inspection, Motion Analysis, and Lighting Consistency Check, among others. By explicitly directing the model’s attention to these diverse and fine-grained visual cues, we significantly enhance its capacity to detect subtle artifacts and inconsistencies commonly found in synthetic videos. Experimental results demonstrate that this prompt-based guidance plays a crucial role in boosting the model’s overall detection performance.

\section{VAE Reconstruction Visualization}
\label{VAE Reconstruction Visualization}

\begin{figure*}[t!]
	\centering
	\includegraphics[width=0.9\linewidth]{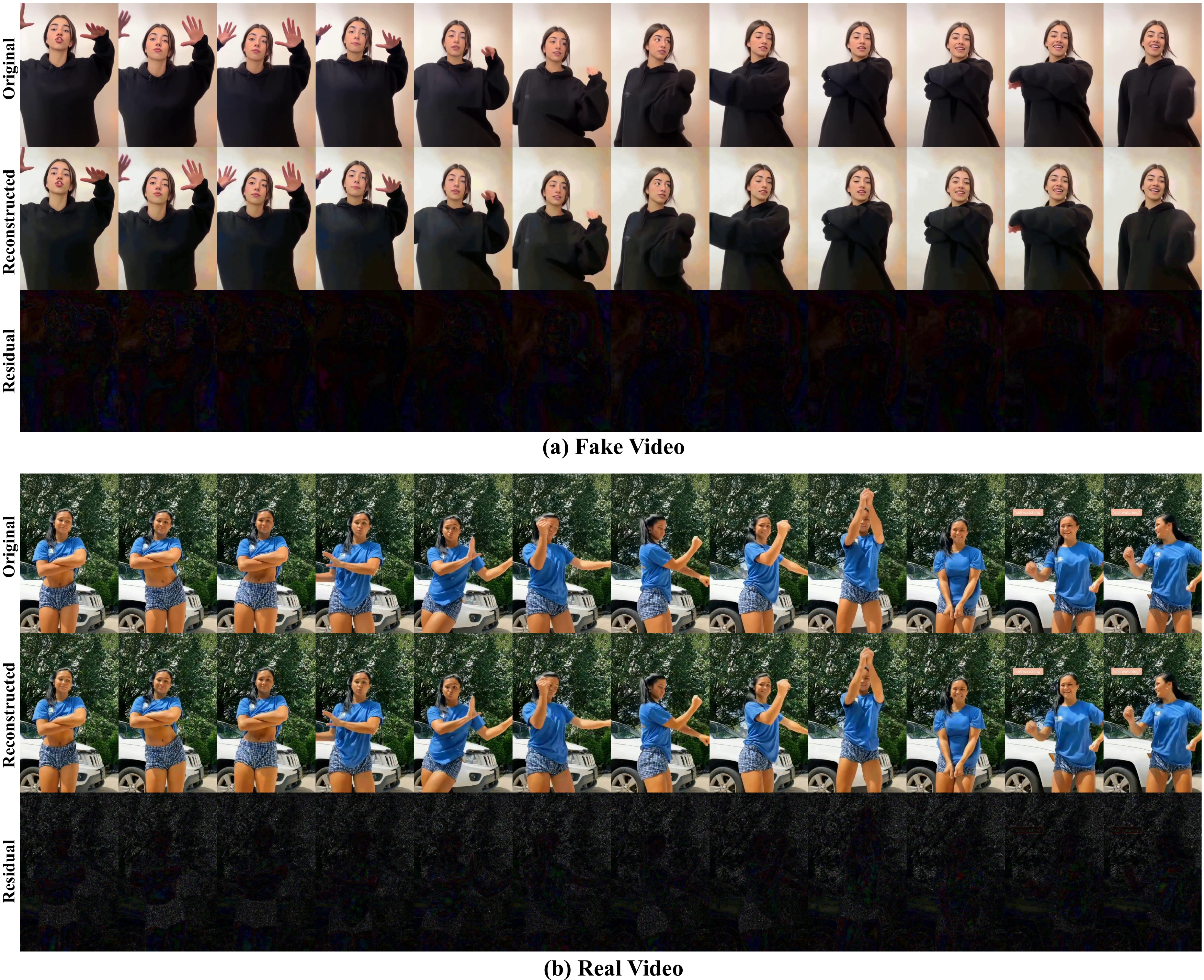}
	\caption{Visualized residual maps, reconstructed videos produced by VQ-VAE, and original videos of the real and fake videos. We observe that the reconstruction residual artifacts of real videos are significantly more noticeable than those of fake videos, which can serve as a basis for forgery detection.}
	\label{img_res_recons}
    \vspace{-16pt}
\end{figure*}

As discussed in Section~\ref{Overall Framework of AvatarShield}, we employed VQ-VAE to amplify generation artifacts in the video.
Diffusion-generated videos tend to align well with the distribution learned by a Variational Autoencoder, so the difference between the original video and its VAE reconstruction is usually small. In contrast, real videos often result in larger residuals after reconstruction.  We visualized the real and fake videos before and after reconstruction, along with their residuals, as shown in Figure~\ref{img_res_recons}. The first example is a fake video where the original, reconstructed, and residual frames are shown from top to bottom. The residuals are barely visible, indicating a small difference. The second example is a real video arranged in the same way, but the residuals are much more noticeable, reflecting a larger reconstruction error. This demonstrates how residuals can help distinguish between real and fake videos.

\section{More Examples}
\label{Appendix More Examples}

\begin{figure*}[t!]
	\centering
	\includegraphics[width=1.0\linewidth]{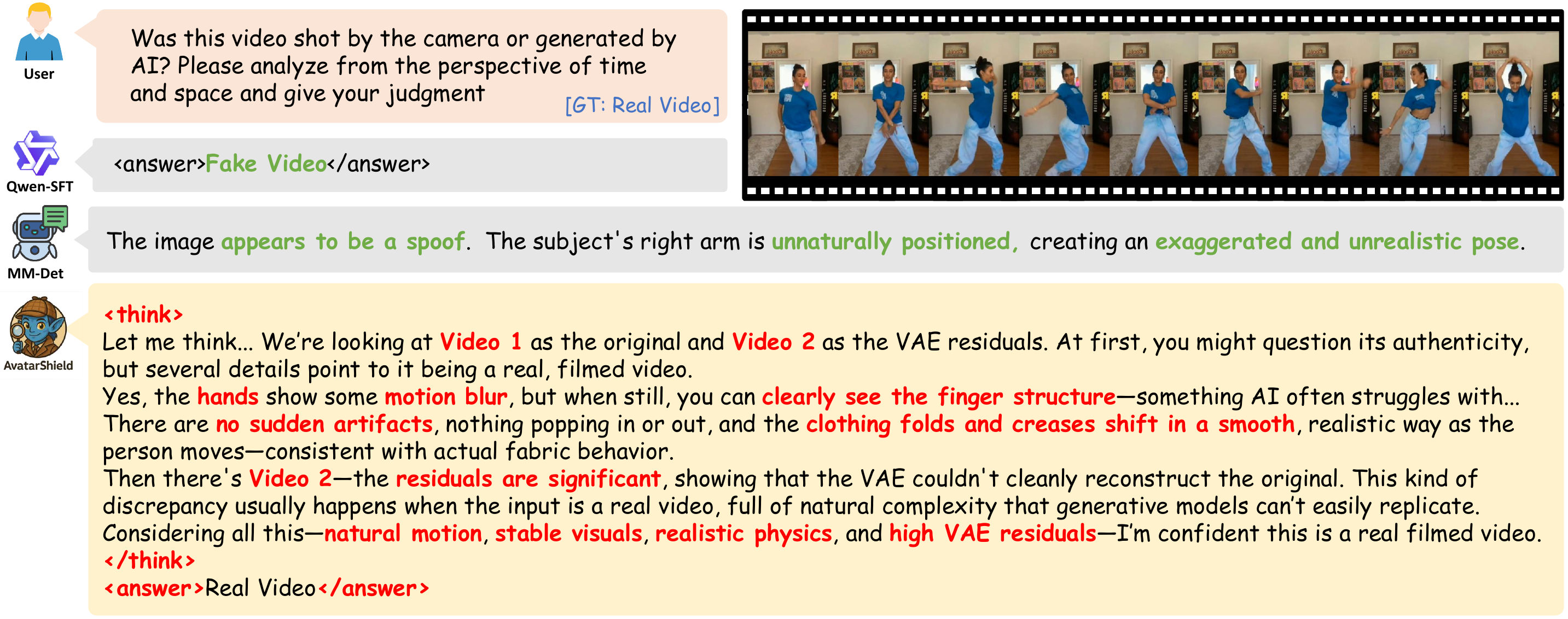}
	\caption{Comparison results between our method, Qwen-SFT, and MM-Det. Our method is the only model that achieves unified precise detection and video interpretation.}
	\label{img_more_compare}
\end{figure*}

\textbf{More qualitative comparisons with LLM-based video forgery detection methods}: As discussed in Section~\ref{In-Domain Detection Results}, we selected additional output samples from LLM-based video forgery detection methods for comparison, as illustrated in Figures~\ref{img_more_compare}.

\begin{figure*}[t!]
	\centering
	\includegraphics[width=1.0\linewidth]{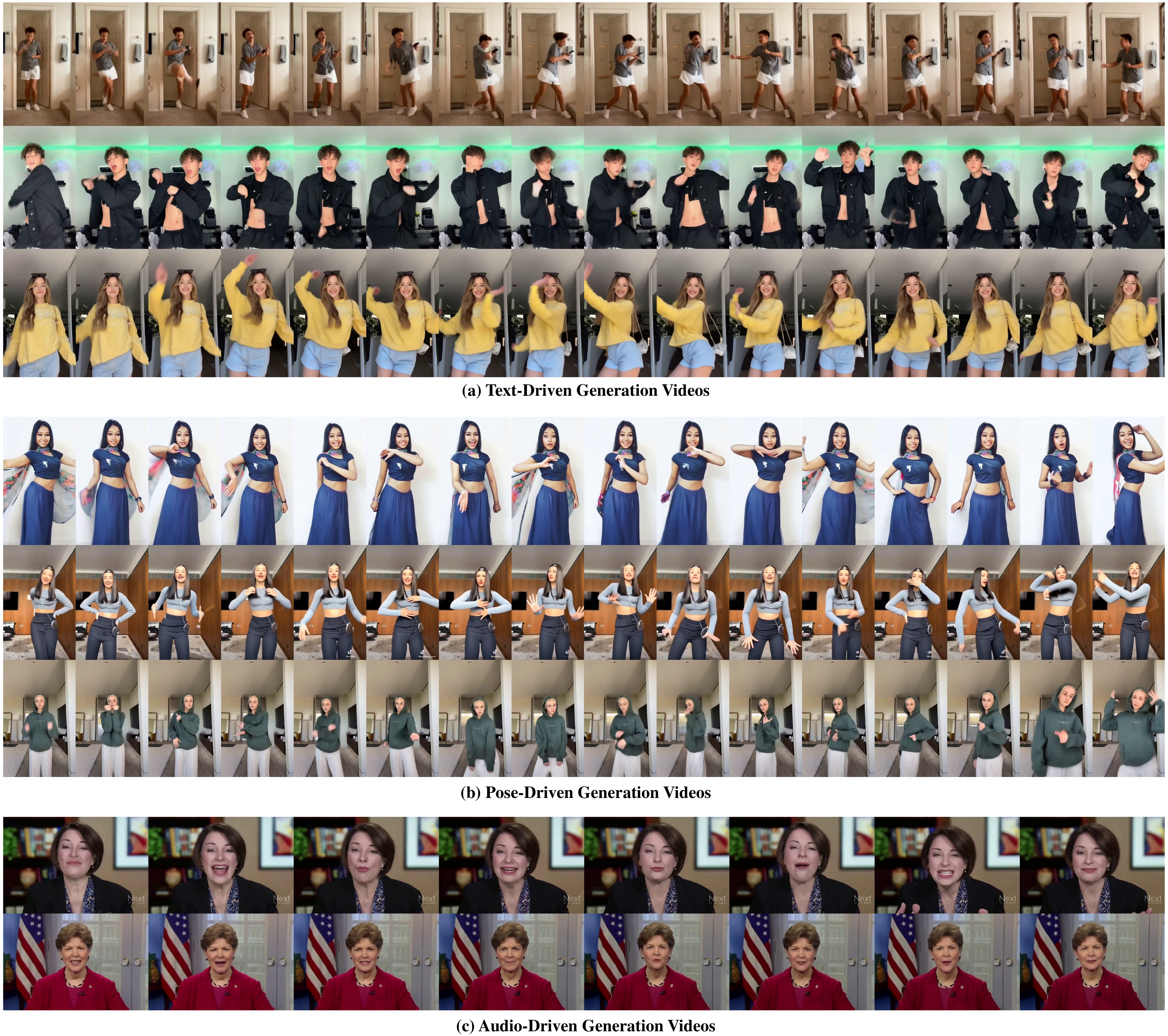}
	\caption{Sampled results of our constructed FakeHumanVid, including text-driven, pose-driven, and audio-driven generation videos.}
	\label{img_more_dataset}
\end{figure*}

\textbf{FakeHumanVid dataset example}: We select some samples from the FakeHumanVid and display them in Figures~\ref{img_more_dataset}.

\begin{figure*}[t!]
	\centering
	\includegraphics[width=0.95\linewidth]{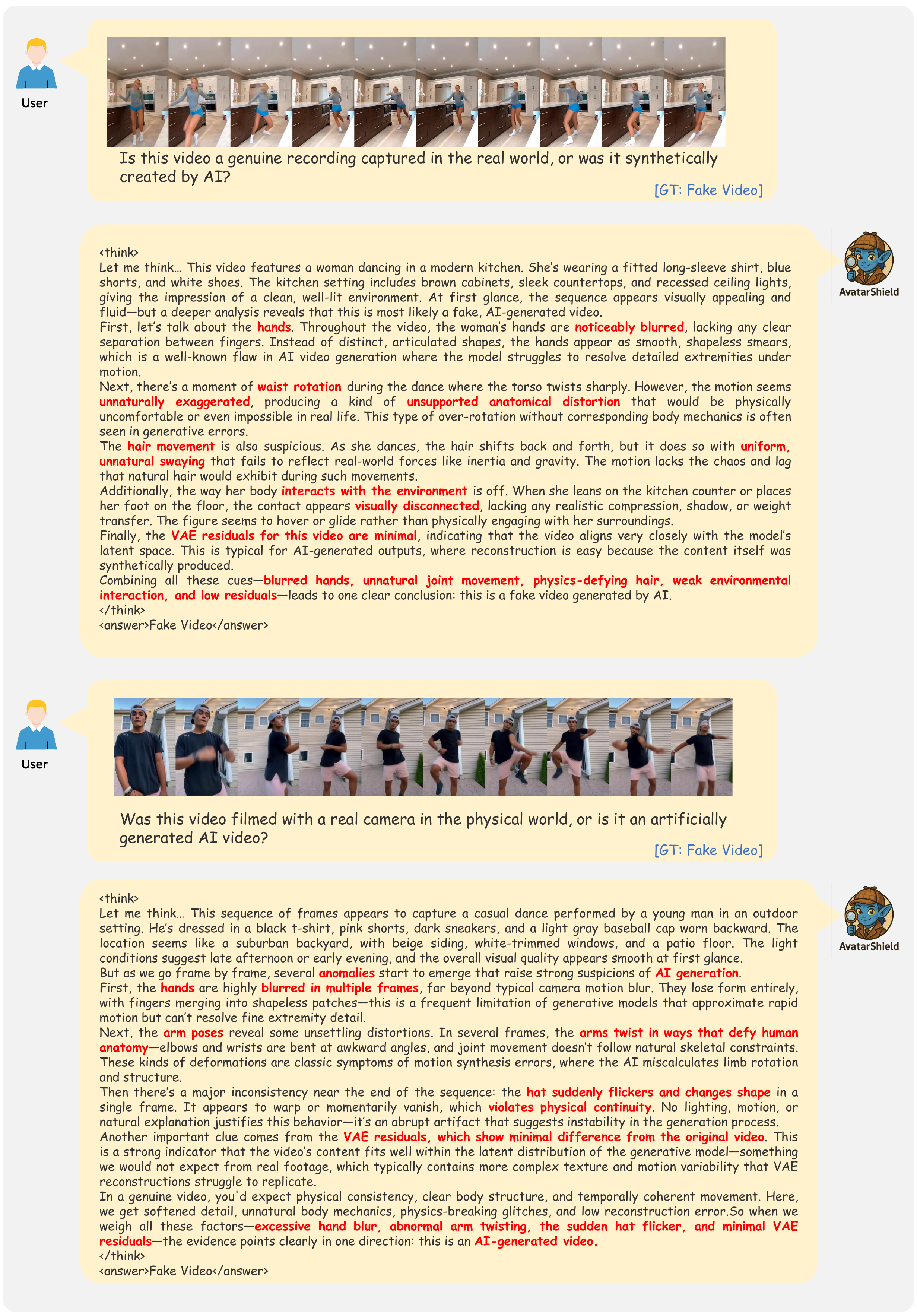}
	\caption{More detection and explanation results of our AvatarShield on text-driven generation videos.}
	\label{img_Example-text}
    \vspace{-16pt}
\end{figure*}

\begin{figure*}[t!]
	\centering
	\includegraphics[width=0.95\linewidth]{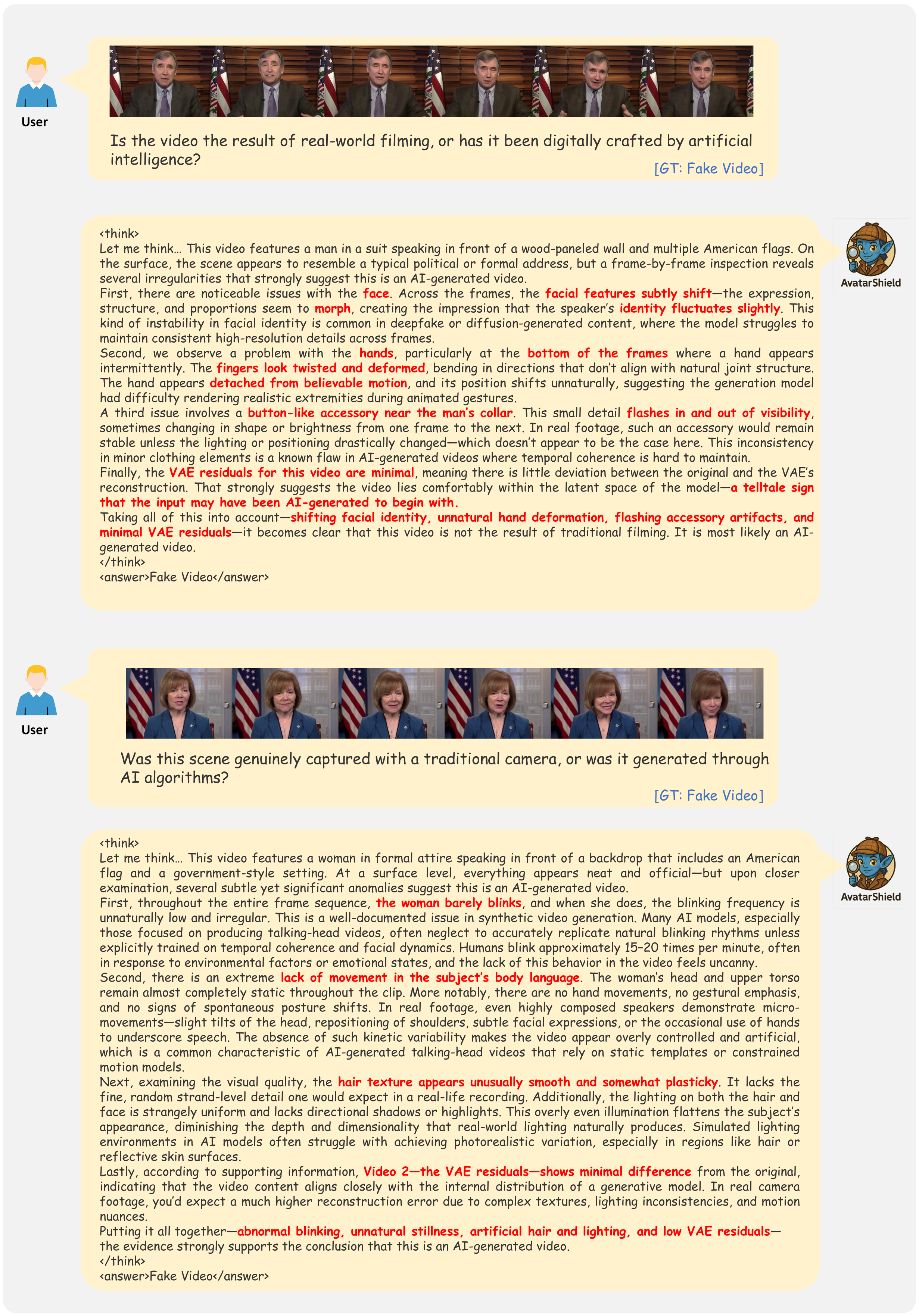}
	\caption{More detection and explanation results of our AvatarShield on audio-driven generation videos.}
	\label{img_Example-audio}
    \vspace{-16pt}
\end{figure*}

\begin{figure*}[t!]
	\centering
	\includegraphics[width=0.95\linewidth]{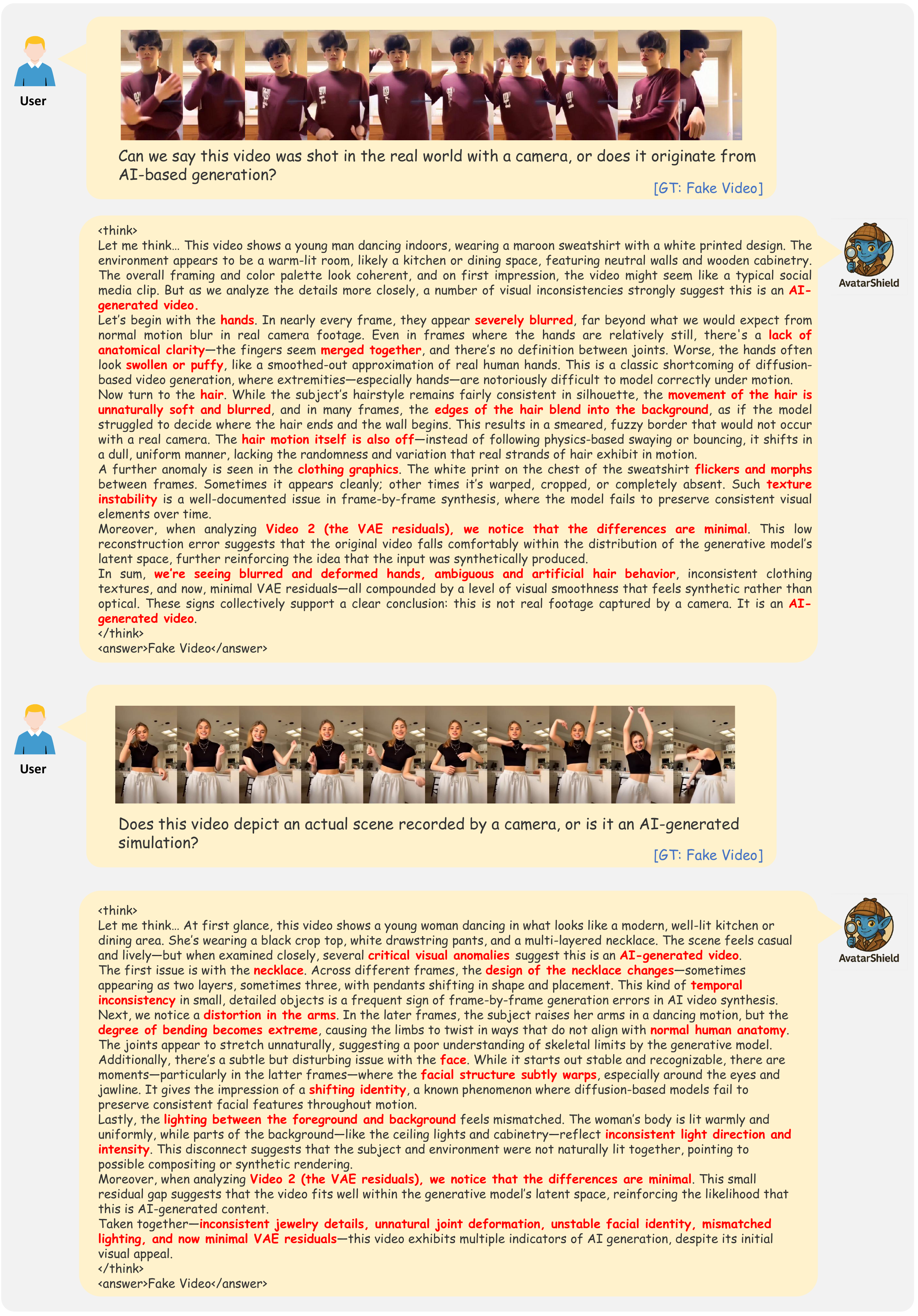}
	\caption{More detection and explanation results of our AvatarShield on pose-driven generation videos.}
	\label{img_Example-pose}
    \vspace{-16pt}
\end{figure*}

\begin{figure*}[t!]
	\centering
	\includegraphics[width=0.9\linewidth]{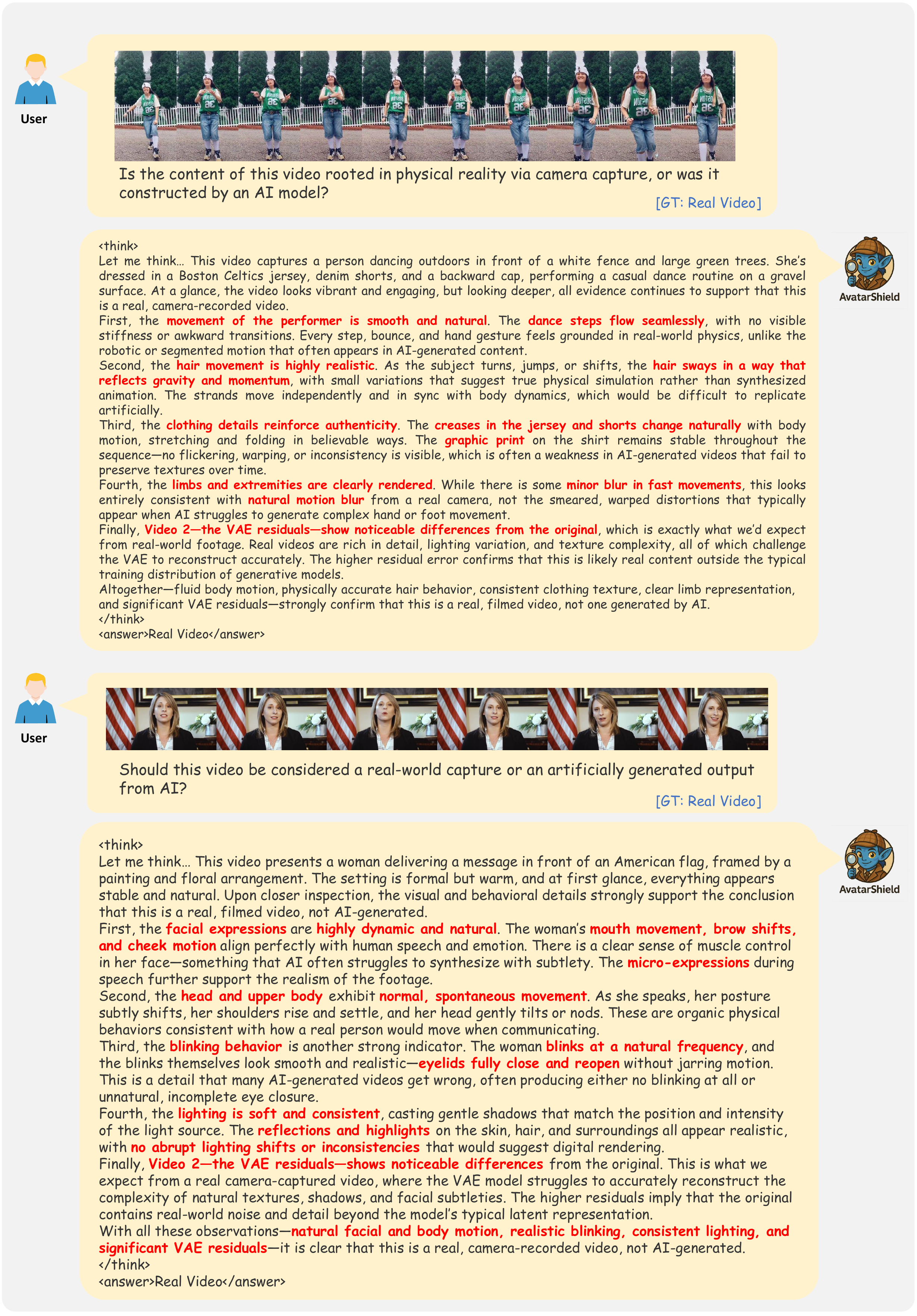}
	\caption{More detection and explanation results of our AvatarShield on real videos.}
	\label{img_Example-au}
    \vspace{-16pt}
\end{figure*}

\textbf{AvatarShield dialog samples}: We selected several dialog samples from AvatarShield's testing on text-driven, audio-driven, pose-driven and authentic datasets, as displayed in Figures~\ref{img_Example-text},~\ref{img_Example-audio},~\ref{img_Example-pose},~\ref{img_Example-au}.

\clearpage

\end{appendices}

\clearpage
\bibliography{sn-bibliography}

\end{document}